\documentclass[letterpaper]{article} 
\usepackage{aaai2027}  
\usepackage[hyphens]{url}  
\usepackage{graphicx} 
\usepackage{natbib}  
\usepackage{caption} 
\usepackage{multirow}
\usepackage{caption}
\usepackage{subcaption}
\usepackage{simpleicons}
\usepackage{algorithm}
\usepackage{algorithmic}
\usepackage{amsmath}
\usepackage{colortbl}
\usepackage{newfloat}
\usepackage{listings}
\DeclareCaptionStyle{ruled}{labelfont=normalfont,labelsep=colon,strut=off} 
\floatstyle{ruled}
\newfloat{listing}{tb}{lst}{}
\floatname{listing}{Listing}

\usepackage{booktabs}

\title{Where and How to Prune: An Empirical Study of Visual Token Pruning for \\ GUI Agent Navigation}
\author {
    Daiqiang Li\textsuperscript{\rm 1,\rm7},
    Zihao Pan\textsuperscript{\rm 2},
    Zeyu Zhang\textsuperscript{\rm 3},
    Xuyang Liu\textsuperscript{\rm 4},
    Shijia Xu\textsuperscript{\rm 5},
    Ronghao Chen\textsuperscript{\rm 6},\\
    Huacan Wang\textsuperscript{\rm 6},
    Honggang Chen\textsuperscript{\rm 7},
    Linfeng Zhang\textsuperscript{\rm 1},
    Zhangquan Chen\textsuperscript{\rm 8},
    Haiyun Jiang\textsuperscript{\rm 1}\corresponding
}
\affiliations {
    \textsuperscript{\rm 1}Shanghai Jiao Tong University~
    \textsuperscript{\rm 2}The University of Hong Kong~
    \textsuperscript{\rm 3}University of California, Berkeley ~ \\
    \textsuperscript{\rm 4}The Hong Kong Polytechnic University~
    \textsuperscript{\rm 5}Chongqing University~ \\
    \textsuperscript{\rm 6}QuantaAlpha~
    \textsuperscript{\rm 7}Sichuan University~
    \textsuperscript{\rm 8}Tsinghua University ~\\
    {\small\textcolor[HTML]{24292F}{\simpleicon{github}}~\url{https://github.com/Daiqiang-Li/HistPrune-GUI}\qquad
    \textcolor[HTML]{FFB300}{\simpleicon{huggingface}}~\url{https://huggingface.co/lidaiqiang/HistPrune-GUI}}
}

\begin{document}
\nocopyright
\maketitle

\begin{abstract}
In recent years, GUI agents have demonstrated strong potential 
in navigation tasks. However, preserving complete historical screenshots introduces substantial computational overhead. This paper investigates how \textbf{token pruning}, a plug-and-play inference acceleration technique, can be effectively applied to GUI agent navigation scenarios. Firstly, we address the question of \textit{where pruning should occur}. We identify a system-level redundancy overlooked by existing methods: as the same screenshot is repeatedly fed into the model across different steps, its ViT encoding is redundantly recomputed each time. We show that its ViT-encoded embeddings can be fully cached and reused across steps, substantially reducing FLOPs while preserving model performance. This finding suggests that inference acceleration efforts should focus on the subsequent Large Language Model (LLM). Building on this, we further address the question of \textit{how to prune} within the LLM, and distill two key insights: \textbf{(i)} from a \textbf{semantic} perspective, the token budget should be balanced between foreground and background regions; \textbf{(ii)} from a \textbf{spatial} perspective, the spatial uniformity of retained tokens should be maintained to preserve the model's global spatial perception. These findings provide practical guidance for the design of inference acceleration and token pruning for GUI agent navigation.
\end{abstract}
\section{Introduction}
\begin{figure}
    \centering
    \includegraphics[width=1.0\linewidth]{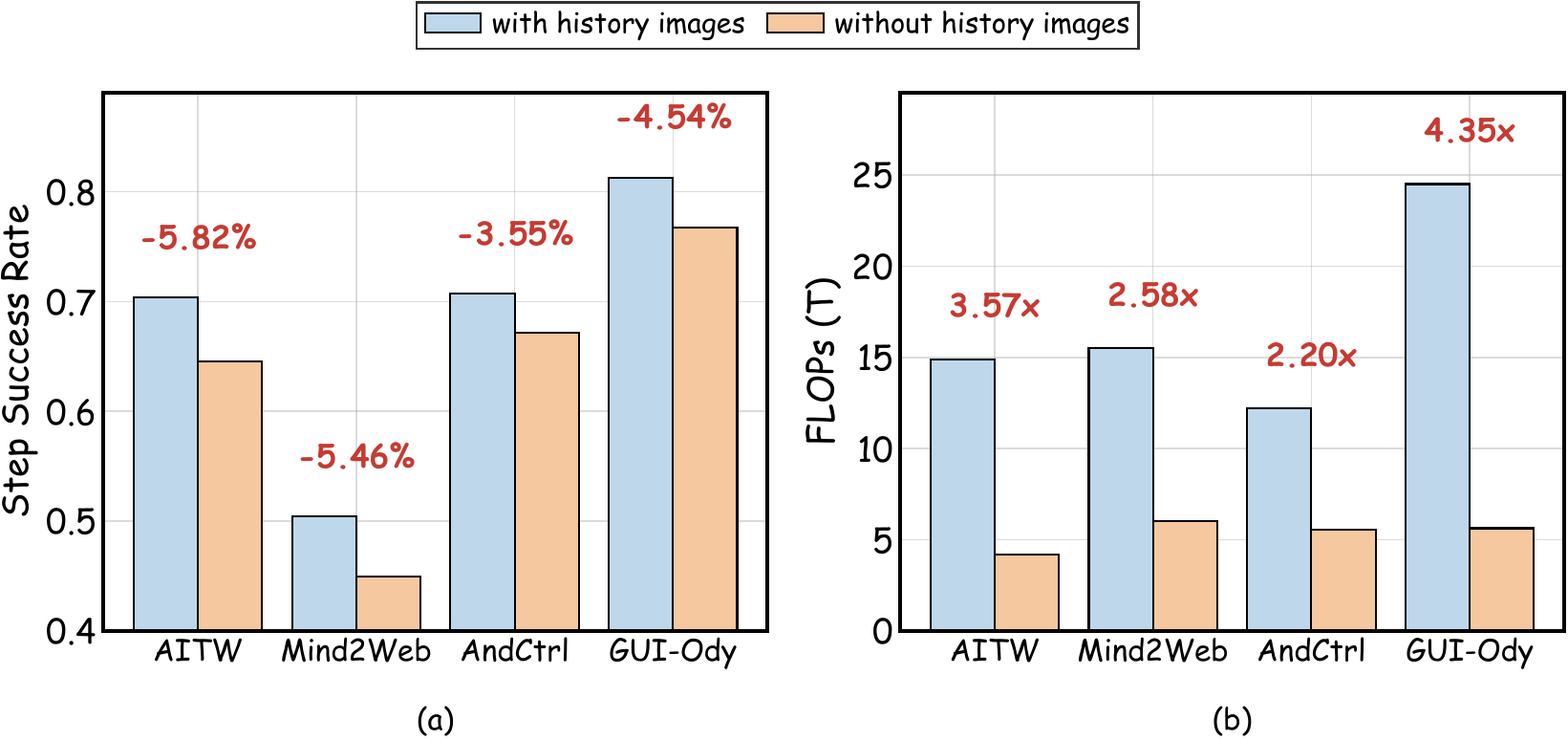}
    \vspace{-1em}
    \caption{\textbf{Performance and computational cost comparison} of Qwen3-VL-2B on GUI agent navigation benchmarks with and without historical screenshots.}
    \label{fig:1}
\end{figure}

In recent years, the development of Multimodal Large Language Models (MLLMs) has substantially advanced research on Graphical User Interface (GUI) agents \cite{visualagent1,visualagent2,GUIG2}. GUI agents take GUI screenshots as input and execute interactions such as clicking, typing, and scrolling in response to natural-language user instructions \cite{agenttask,UI-VENUS}. In practice, the agent's input typically includes the current GUI screenshot, a sequence of historical screenshots and historical actions, and the user's instruction~\cite{ShowUI}. As shown in Figure \ref{fig:1}(a), removing historical screenshots from the input causes a marked drop in the model's step success rate, indicating that historical visual context is beneficial for GUI reasoning and should not be simply discarded. However, as illustrated in Figure \ref{fig:1}(b), preserving historical screenshots substantially increases computational cost. Therefore, designing effective compression strategies for historical screenshots is a meaningful research problem.

Several prior works have attempted to address this problem. SimpAgent \cite{simpagent} discards all historical screenshot tokens entirely and adopts self-distillation to maintain model performance. OdysseyAgent \cite{guiodyssey} employs a history resampler module to compress historical information. However, both methods require retraining. These limitations motivate a training-free and plug-and-play
alternative for compressing historical visual context. Token pruning,
which reduces computation by removing redundant visual tokens at inference time, is a natural candidate.
However, existing token pruning methods are primarily developed for
visual reasoning over natural images or videos, and it remains unclear
whether their design choices transfer to GUI agent navigation. 
To bridge this gap, we systematically investigate where token pruning
should be applied and how historical visual tokens should be retained
in GUI agents.

We first address the question of \textit{where to prune historical screenshots in GUI agent navigation}. In GUI agent navigation tasks, the same screenshot is fed into the model more than once. Since the ViT encodes each image independently, the same screenshot produces identical ViT features regardless of whether it is provided as the current or historical input. Therefore, the features extracted during a screenshot's first pass through the ViT can be fully cached and directly reused in subsequent steps, substantially reducing FLOPs while preserving model performance. To the best of our knowledge, existing works on GUI agent navigation have not recognized this system-level redundancy. Furthermore, we compare against approaches that compress historical screenshots prior to ViT encoding, including patch removal and resolution reduction. We find that both consistently introduce non-trivial performance degradation, suggesting that compressing historical screenshots before ViT encoding is not an 
ideal acceleration path. These findings together indicate that  \textbf{inference acceleration efforts should focus on the subsequent Large Language Model}.

Building on this, we further address the question of \textit{how to prune within the LLM}, and distill two key insights. From a \textbf{semantic perspective}, GUI screenshots typically contain large homogeneous background regions. Intuitively, such backgrounds may seem disposable. However, our empirical analysis shows that this assumption does not hold. Therefore, when designing token pruning methods, the token budget should be carefully balanced between foreground and background regions. From a \textbf{spatial perspective}, token pruning alters the composition of the original token sequence, thereby affecting the model's spatial understanding capability. Through an analysis of the RoPE positional encoding mechanism, we find that spatially concentrated pruning weakens the model's ability to infer global token positions. Our empirical results further show that random pruning naturally maintains a spatially uniform distribution, with retained tokens acting as spatial anchors that provide sufficient positional references, thereby achieving competitive performance. Therefore, the spatial uniformity of retained tokens should be preserved when designing token pruning methods.

In summary, the contributions are as follows:
\begin{itemize}
\setlength{\itemsep}{0pt}
\setlength{\parskip}{0pt}
\item \textbf{Pruning Location}: We identify a previously overlooked system-level redundancy in GUI agent navigation: the same screenshot is repeatedly fed into the model across different steps, yet its ViT features remain identical. We propose a cache-and-reuse mechanism that substantially reduces FLOPs with negligible performance loss. Furthermore, we show that compressing historical screenshots before ViT encoding consistently leads to non-trivial performance degradation. These findings together demonstrate that inference acceleration and token pruning efforts should focus on the subsequent LLM.
\item \textbf{Pruning Strategy}: We distill two key principles for token pruning of historical screenshots within the LLM: \textbf{(i)} from a semantic perspective, the token budget should be appropriately balanced between foreground and background regions; 
\textbf{(ii)} from a spatial perspective, the spatial uniformity of retained tokens should be carefully maintained to preserve the model's global spatial perception.
\end{itemize}
By systematically addressing the two fundamental questions of where to prune and how to prune, we uncover a series of valuable findings. We hope these findings provide useful insights and practical guidance for future research on efficient inference and token pruning in GUI agent navigation.
\section{Related Work}
\subsection{Inference Acceleration for GUI Agent Navigation}
Incorporating historical screenshots into GUI agent navigation tasks substantially increases FLOPs. Several prior works have recognized this problem and attempted to compress historical screenshots. SimpAgent completely removes all historical screenshot tokens during training and adopts a self-distillation approach, using KL divergence constraints to train the model such that inference-time performance is maintained with negligible degradation. OdysseyAgent designs a history resampler module to compress historical information. However, its query vectors require dedicated training. Both methods rely on additional model training and lack plug-and-play capability. GUIPruner~\cite{GUIPruner} proposes a training-free inference acceleration method that reduces the resolution of historical screenshots prior to feeding them into the model. However, this pre-encoding resolution reduction disrupts the model's spatial perception of the interface, leading to a notable drop in performance. Furthermore, none of the above methods have recognized an inherent characteristic of GUI navigation tasks: the same screenshot is repeatedly fed into the model across different steps, meaning that the ViT-encoded features extracted during a screenshot's first pass can be cached and reused in subsequent steps. This issue will be discussed in detail in this paper.

\subsection{Token Pruning for MLLMs}
In recent years, token pruning methods for Multimodal Large Language Models have attracted wide attention~\cite{liu2025shifting,lin2026v-cast}. Existing approaches can be broadly categorized into two groups: importance-based methods, such as FastV~\cite{FastV}, SparseVLM~\cite{sparsevlm}, PDrop~\cite{PDrop}, and GlobalCom$^2$~\cite{global}; and redundancy-based methods, such as DART~\cite{DART}, DivPrune~\cite{Divprune}, VidCom$^2$~\cite{liu2025video}, and V$^2$Drop~\cite{chen2025variation}. However, these methods are primarily designed for visual reasoning and scene understanding in natural images and videos, and are not tailored for GUI agent scenarios. 

Moreover, existing token pruning methods operate at different locations within the model, either within the ViT encoder~\cite{IPCV,STC,han2026ficoco} or within the subsequent LLM~\cite{FastV,sparsevlm,MixKV}. However, which pruning location is more suitable for GUI agent navigation has not been investigated. To address this gap, this paper investigates token pruning for GUI agent navigation.
\section{Experiment Setup}

\noindent\textbf{Benchmarks.}
We conduct empirical studies on four widely used GUI agent benchmarks: AITW~\cite{AITW}, Mind2Web~\cite{Mind2Web}, AndroidControl~\cite{AndroidControl} (AndCtrl), and GUI-Odyssey~\cite{guiodyssey} (GUI-Ody). For Mind2Web, we report results under the cross-domain (M-d), cross-task (M-t), and cross-website (M-w) settings. Detailed descriptions are provided in Appendix~\ref{app:dataset_details}.

\noindent\textbf{Models.}
We conduct experiments using two representative multimodal large language models: Qwen2.5-VL-3B~\cite{Qwen2.5VL} and Qwen3-VL-2B~\cite{Qwen3VL}. Notably, Qwen3-VL additionally incorporates DeepStack~\cite{deepstack}, a mechanism that injects intermediate ViT feature representations from multiple layers into the LLM layers via lightweight residual connections, enabling richer vision-language alignment. 

\noindent\textbf{Historical Context Setting.}
Following SimpAgent~\cite{simpagent}, we retain up to 4 historical steps for AITW and GUI-Odyssey, and 2 for Mind2Web and AndroidControl.

\noindent\textbf{Training and Evaluation Details.}
We first perform supervised fine-tuning on all four benchmarks to establish strong baselines. Details of the training configuration, hardware setup, FLOPs measurement, GPU peak memory measurement, and evaluation metrics are provided in Appendix~\ref{app:training_eval_details}.

\noindent\textbf{Pruning Setting.}
In all token pruning experiments, we only prune visual tokens from historical screenshots, while keeping the current screenshot, historical actions, and user instructions unchanged. The retention ratio $R$ denotes the fraction of visual tokens retained from each historical screenshot. Pruning is applied at the 4th LLM layer, with the rationale provided in the Appendix~\ref{app:pruning_layer}.
\section{Problem Statement}
\label{sec:formulation}

\begin{figure*}
    \centering
    \includegraphics[width=0.9\linewidth]{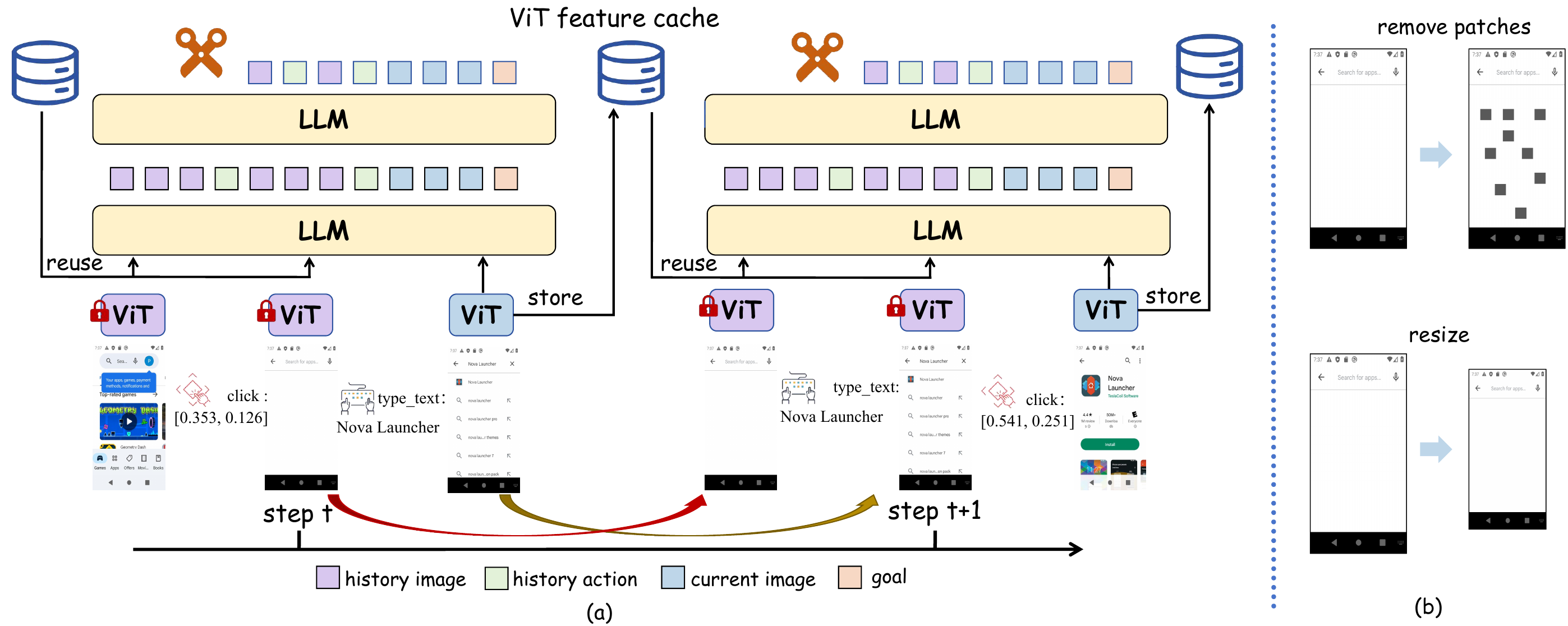}
    \caption{(a) \textbf{The ViT feature caching mechanism}, which caches and reuses ViT-encoded features of repeated screenshots across steps. For Qwen3-VL with DeepStack, the intermediate ViT features are cached, but are omitted from the illustration for clarity. (b) Two approaches that compress historical screenshots before ViT encoding: \textbf{patch removal and resolution reduction}.}
    \label{fig:2}
\end{figure*}

GUI agent navigation can be naturally modeled as a sequential decision-making problem. At each timestep $t$, the agent predicts the next action $a_t$ based on the current screenshot $o_t$, the historical screenshot sequence $I_t$, the past action sequence $A_t$, and the task goal $G$:
\begin{equation}
    a_t = \pi_\theta(o_t, I_t, A_t, G)
\end{equation}
where $\pi_\theta$ denotes the GUI agent parameterized by $\theta$, trained to minimize the negative log-likelihood:
\begin{equation}
    L = -\sum_t \log P(a_t \mid o_t, I_t, A_t, G)
\end{equation}
Mainstream methods typically truncate the input history, retaining only the most recent $\tau$ steps:
\begin{equation}
    I_t = \{o_{t-\tau},\mathinner{{\ldotp}{\ldotp}}, o_{t-1}\},A_t = \{a_{t-\tau},\mathinner{{\ldotp}{\ldotp}}, a_{t-1}\}
\end{equation}
However, even when the history is truncated to the most recent $\tau$ steps, the retained historical screenshots still introduce substantial computational overhead. This motivates the present work to investigate token pruning mechanisms for GUI agents, aiming to reduce computational cost while preserving model performance.

\section{Pruning Location}
\label{sec:pruning_location}
As shown in Figure~\ref{fig:1}, the additional FLOPs in GUI agent navigation mainly come from incorporating historical screenshots. Meanwhile, the current screenshot provides the most direct visual evidence for action prediction, and pruning it leads to substantial performance degradation, as analyzed in Appendix~\ref{app:current_screenshot_pruning}. Therefore, we focus on pruning visual tokens from historical screenshots while keeping the current screenshot unchanged throughout our experiments.

Revisiting the formulation in the Problem Statement section, at timestep $t$, the model retains the most recent $\tau$ historical screenshots, namely:
\begin{equation}
    I_t = \{o_{t-\tau}, \ldots, o_{t-1}\}
\end{equation}
This means that after a screenshot $o_s$ first appears as the current observation, it subsequently reappears as a historical screenshot for $\tau$ consecutive timesteps, or until the task terminates. Following the convention of existing methods, $o_s$ is re-encoded by the vision encoder at every timestep. However, the vision encoder processes each image independently, without dependence on the current timestep or surrounding context. For the same image, the encoder always produces identical ViT features. We therefore propose to cache the feature vectors of each screenshot upon its first encoding, and directly reuse them in subsequent timesteps when the screenshot reappears as a historical observation, without any recomputation. The proposed 
mechanism is illustrated in Figure~\ref{fig:2}(a). For Qwen3-VL, which employs the DeepStack technique, we additionally cache the intermediate ViT features that are subsequently injected into the LLM. As shown in Table~\ref{tab:cache_perf}, introducing this cache-and-reuse mechanism yields nearly identical performance to the original approach across all benchmarks. The marginal differences are not caused by the caching mechanism itself. Instead, they stem from minor numerical discrepancies in floating-point computation. As shown in Table~\ref{tab:flops}, this mechanism substantially reduces 
FLOPs by 31.5\% and accelerates inference, achieving an 11.8\% reduction in wall-clock time and a 67\% reduction in per-step encoder latency.
Furthermore, since the cache is cleared at the end of each episode, 
it introduces negligible additional GPU peak memory overhead, with 
only a 1.5\% increase compared to the original approach.

\definecolor{greenrightcolor}{RGB}{0,144,81} 
\definecolor{redrightcolor}{RGB}{255,0,0} 

\newcommand{\downtinya}[1]{{{\tiny{#1}}}}
\newcommand{\uptiny}[1]{{{\tiny{#1}}}}

\begin{table*}[h]
\centering
\footnotesize
\resizebox{\textwidth}{!}{
\begin{tabular}{l c c c c c c}
\toprule
\multirow{2}{*}{\textbf{Strategies}} 
& \textbf{FLOPs$\downarrow$} 
& \textbf{Memory$\downarrow$} 
& \textbf{Wall-clock time$\downarrow$} 
& \textbf{Encoder latency$\downarrow$} 
& \textbf{LLM prefill latency$\downarrow$} 
& \multirow{2}{*}{\textbf{Performance$\uparrow$}} \\
& \textbf{(T)} 
& \textbf{(MB)} 
& \textbf{(min)} 
& \textbf{(ms/step)} 
& \textbf{(ms/step)} 
& \\
\midrule
With HI            
& 14.9 
& 5067 
& 108.54 
& 212.08 
& 193.05 
& 0.7035 \\

\midrule
Without HI           
& 4.17 \downtinya{($\downarrow$72.0\%)}
& 4289 \downtinya{($\downarrow$15.4\%)} 
& 76.62 \downtinya{($\downarrow$29.4\%)} 
& 67.90 \downtinya{($\downarrow$68.0\%)}  
& 151.70 \downtinya{($\downarrow$21.4\%)} 
& 0.6453 \downtinya{($\downarrow$8.27\%)} \\

ViT Cache            
& 10.2 \downtinya{($\downarrow$31.5\%)} 
& 5144 \uptiny{($\uparrow$1.52\%)} 
& 95.70 \downtinya{($\downarrow$11.8\%)} 
& 69.82 \downtinya{($\downarrow$67.1\%)} 
& 202.54 \uptiny{($\uparrow$4.92\%)} 
& 0.7030 \downtinya{($\downarrow$0.07\%)} \\

Resize ($\times$0.8) 
& 10.9 \downtinya{($\downarrow$26.8\%)} 
& 4795 \downtinya{($\downarrow$5.37\%)} 
& 97.69 \downtinya{($\downarrow$10.0\%)} 
& 151.73 \downtinya{($\downarrow$28.5\%)} 
& 144.63 \downtinya{($\downarrow$25.1\%)} 
& 0.6965 \downtinya{($\downarrow$1.00\%)} \\

Patch Removal        
& 11.0 \downtinya{($\downarrow$26.2\%)} 
& 4606 \downtinya{($\downarrow$9.10\%)} 
& 98.02 \downtinya{($\downarrow$9.69\%)} 
& 162.11 \downtinya{($\downarrow$23.6\%)} 
& 149.47 \downtinya{($\downarrow$22.6\%)} 
& 0.6859 \downtinya{($\downarrow$2.50\%)} \\

\bottomrule
\end{tabular}
} 
\caption{Comparison of FLOPs, GPU peak memory, wall-clock time, encoder latency, LLM prefill latency, and step success rate for different strategies on the AITW dataset, based on Qwen3-VL-2B. HI denotes historical images. Parenthesized values report relative changes with respect to With HI.}
\label{tab:flops}
\end{table*}

\begin{table}[h]
\centering

\resizebox{\columnwidth}{!}{%
\begin{tabular}{lcccccc}
\toprule
& \textbf{AITW} & \textbf{M-d} & \textbf{M-t} & \textbf{M-w}
& \textbf{AndCtrl} & \textbf{GUI-Ody} \\
\midrule
Original & 0.7035 & 0.4863 & 0.5363 & 0.4858 & 0.7069 & 0.8128 \\
ViT Cache & 0.7030 & 0.4884 & 0.5363 & 0.4829 & 0.7061 & 0.8136 \\
\bottomrule
\end{tabular}
}
\caption{\textbf{Performance comparison} of Qwen3-VL-2B with and without the proposed ViT cache-and-reuse mechanism on four benchmarks.}
\label{tab:cache_perf}
\end{table}

While the ViT cache-and-reuse mechanism avoids redundant re-encoding, one may naturally ask: \textit{Could we go further by compressing historical screenshots before they are passed to the ViT}? As shown in Figure~\ref{fig:2}(b), one straightforward strategy is to directly remove a subset of patches. However, GUI agent navigation demands strong spatial awareness of image content, and directly removing patches disrupts the spatial structure, leading to a notable performance drop. Specifically, we randomly remove 36\% of patches from each historical screenshot prior to ViT encoding. As shown in Table~\ref{tab:flops}, despite achieving FLOPs and inference speedup comparable to the ViT caching mechanism, patch removal leads to an absolute drop in step success rate of 1.76\%, substantially larger than the 0.05\% drop of ViT caching. 

A more moderate alternative is to reduce the resolution of historical screenshots. As shown in 
Table~\ref{tab:flops}, on the AITW dataset, compressing the height and width of historical screenshots to $0.8\times$ their original size reduces FLOPs to a level comparable to that of the ViT caching mechanism, yet introduces a step success rate absolute drop of 0.70\%, considerably larger than the negligible 0.05\% drop of ViT caching. Furthermore, as shown in Table~\ref{tab:resolution}, this degradation is not limited to AITW. Resolution compression generally leads to varying degrees of performance loss across all benchmarks.

Based on the above analysis, the focus of this paper is neither on processing images prior to the ViT nor on pruning within the ViT, but rather on the large language model that follows. We further analyze LLM KV-cache reuse in Appendix~\ref{app:llm-kv-cache-analysis}, showing that it is only safe for exact shared prefixes and cannot be generally applied to historical GUI segments after the history window slides.

\begin{table}[t]
\centering
\resizebox{\columnwidth}{!}{%
\begin{tabular}{lcccccc}
\toprule
\multicolumn{7}{c}{\textit{\textbf{Qwen2.5-VL-3B}}} \\
\midrule
\textbf{Res} & \textbf{AITW} & \textbf{M-d} & \textbf{M-t} & \textbf{M-w} & \textbf{AndCtrl} & \textbf{GUI-Ody} \\
\midrule
$\times$1.0 & 0.6806 & 0.4889 & 0.5580 & 0.4697 & 0.7113 & 0.7790 \\
$\times$0.9 & 0.6794 & 0.4865 & 0.5482 & 0.4710 & 0.7110 & 0.7730 \\
$\times$0.8 & 0.6806 & 0.4818 & 0.5382 & 0.4687 & 0.7094 & 0.7670 \\
\midrule
\multicolumn{7}{c}{\textit{\textbf{Qwen3-VL-2B}}} \\
\midrule
\textbf{Res} & \textbf{AITW} & \textbf{M-d} & \textbf{M-t} & \textbf{M-w} 
& \textbf{AndCtrl} & \textbf{GUI-Ody} \\
\midrule
$\times$1.0 & 0.7035 & 0.4863 & 0.5363 & 0.4858 & 0.7069 & 0.8128 \\
$\times$0.9 & 0.6976 & 0.4869 & 0.5310 & 0.4814 & 0.7045 & 0.8104 \\
$\times$0.8 & 0.6965 & 0.4837 & 0.5295 & 0.4726 & 0.7036 & 0.8091 \\
\bottomrule
\end{tabular}%
}
\caption{\textbf{Performance under different resolution compression ratios} for Qwen2.5-VL-3B and Qwen3-VL-2B.}
\label{tab:resolution}
\end{table}
\section{Semantic Perspective: Balance Foreground and Background}

\begin{figure*}
    \centering
    \includegraphics[width=0.9\linewidth]{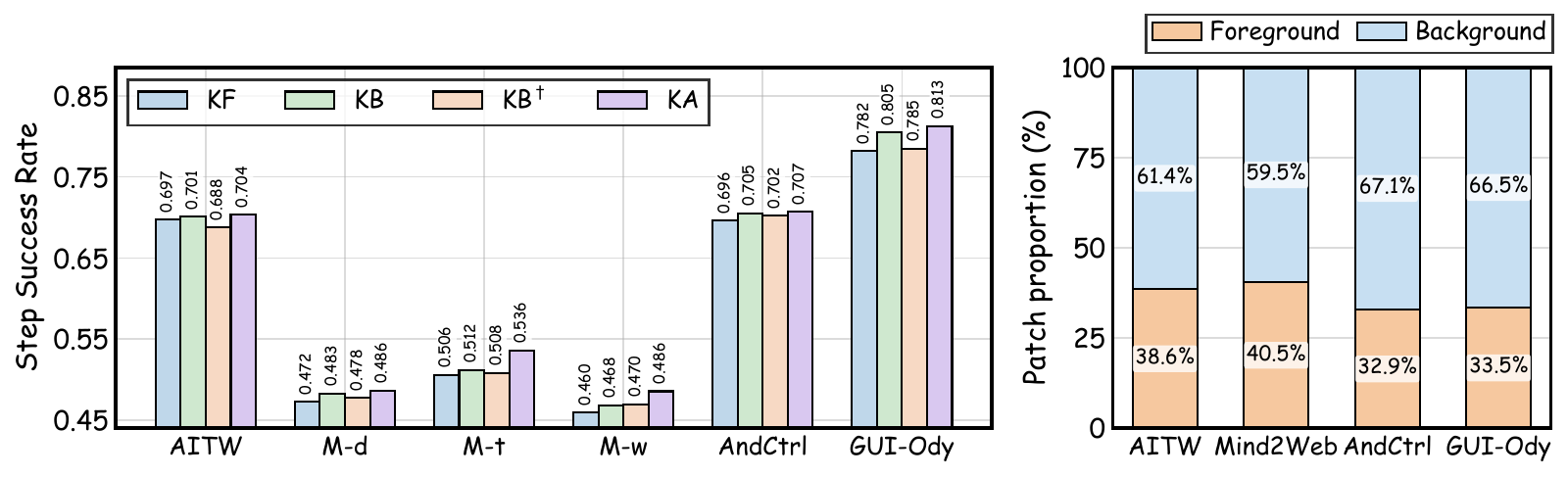}
    \caption{Left: Step success rate of different token retention 
strategies (KF: keep foreground only; KB: keep background only; 
KB$^\dagger$: keep background subsampled to match KF budget; KA: 
keep all) across four benchmarks. Right: Proportion of foreground 
and background patches across four GUI agent benchmarks, where 
background patches account for 63.63\% on average.}
    \label{fig:3}
\end{figure*}

\begin{figure}[t]
    \centering
    \includegraphics[width=1.0\linewidth]{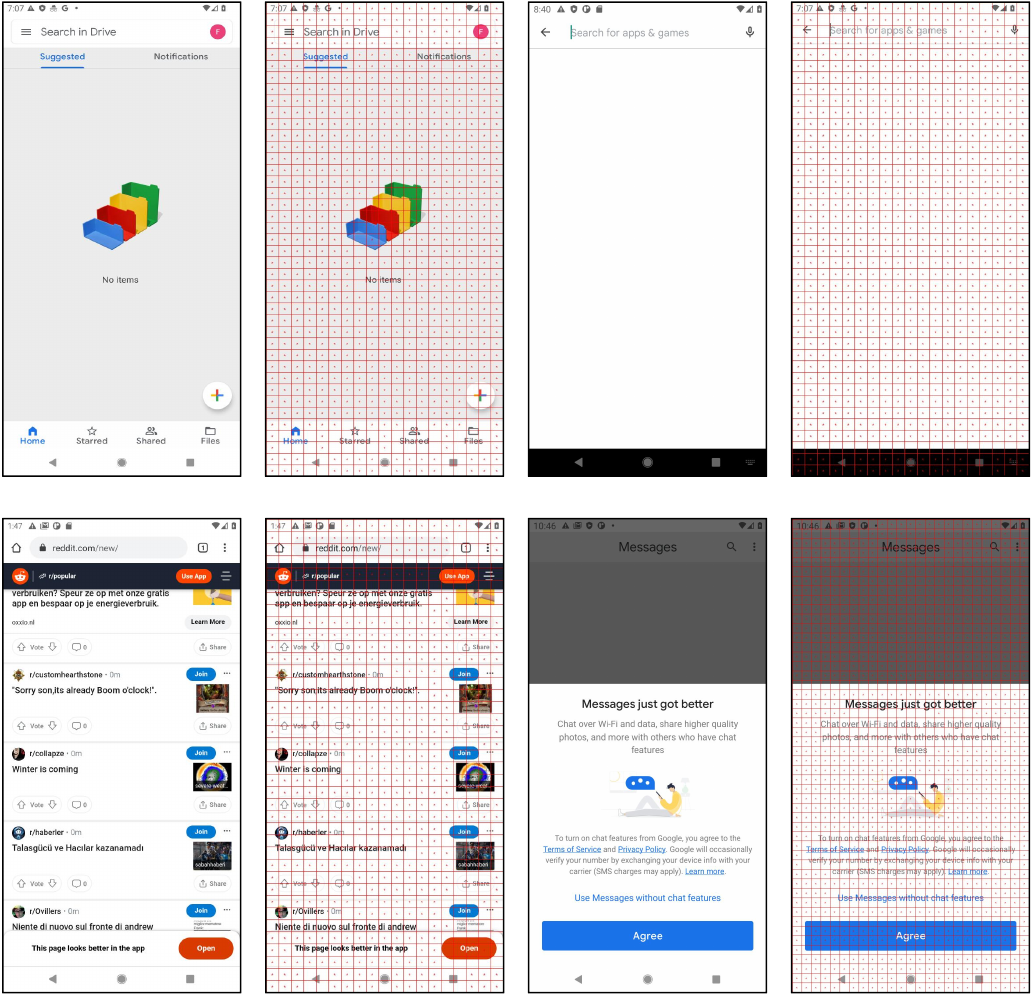}
    \caption{Illustration of the edge-based foreground–background partition on a GUI screenshot.}
    \label{fig:4}
\end{figure}

Unlike natural images, GUI screenshots are characterized by the pervasive presence of large homogeneous background regions. In GUI interfaces, semantically meaningful elements, such as text, icons, buttons, and component boundaries, are typically associated with local edges or high-frequency visual structures. In contrast, regions with little or no edge response are usually homogeneous background areas and often contain limited explicit semantic information. Motivated by this observation, we use edge information as a lightweight proxy for foreground-background partitioning. To systematically analyze this property, we first partition each screenshot into visual patches using the same patch strategy adopted by the MLLM, ensuring consistency with the model's input pipeline. We then apply Sobel edge detection to determine whether each patch contains significant edge information. Specifically, the input image is first converted to grayscale and smoothed with a Gaussian filter (kernel size $5\times5$, $\sigma=1.0$) to suppress noise. The horizontal and vertical Sobel responses $S_x$ and $S_y$ are then computed by convolving the smoothed image with standard $3\times3$ Sobel kernels in the horizontal and vertical directions, respectively, and the gradient magnitude is calculated as:
\begin{equation}
    M(i,j) = \sqrt{S_x(i,j)^2 + S_y(i,j)^2}
\end{equation}
The magnitude map is then normalized to $[0, 255]$:
\begin{equation}
    \hat{M}(i,j) = \frac{M(i,j)}{\max_{i,j} M(i,j)} \times 255
\end{equation}
For each patch, we compute the ratio of pixels whose normalized 
magnitude exceeds an edge threshold $\tau_e$:
\begin{equation}
    r = \frac{1}{ N^2} \sum_{(i,j) \in \text{patch}} \mathbf{1}\left[\hat{M}(i,j) \geq \tau_e\right]
\end{equation}
where $N \times N$ denotes the spatial resolution of each patch 
in the original image, as determined by the MLLM's patch strategy. 
A patch is labeled as foreground if $r \geq \tau_r$, and as background otherwise. In this paper, $\tau_e = 50$ and $\tau_r = 0.01$. As shown in Figure~\ref{fig:4}, this approach enables effective segmentation of historical screenshots. Additional segmentation visualizations are provided in Appendix~\ref{app:segmentation_visualization}.

Based on this partition, we compute the proportion of foreground and background patches across four  GUI agent benchmarks. As shown in Figure~\ref{fig:3} (right), background patches account for a substantial proportion across all datasets, with an average ratio of approximately 63.63\%. An intuitive hypothesis is that background regions are semantically impoverished and can be safely discarded. To verify this hypothesis, we design two ablation conditions: Keep Foreground (KF), which retains only foreground patches; and Keep Background (KB), which retains only background patches. As shown in Figure~\ref{fig:3} (left), the results reveal a counter-intuitive finding: retaining background patches alone yields even better performance than retaining foreground patches alone.

It is worth noting that background patches naturally outnumber foreground patches, which may make the above comparison unfair due to the token budget imbalance. To address this, we conduct an additional controlled experiment, KB$^\dagger$, where background patches are subsampled to match the token count of KF. The detailed construction of KB$^\dagger$ is provided in Appendix~\ref{app:kb_dagger}. As shown in Figure~\ref{fig:3}, even under this controlled budget, KB$^\dagger$ still outperforms KF on most benchmarks.

We offer two explanations for this phenomenon. First, although background regions are commonly regarded as "low semantics", their changes across timesteps encode environment and interface state transition information, such as shifts in window positions, layout changes, which can provide important decision-making context for the agent. Second, the agent's decisions rely primarily on the current screenshot, with historical screenshots serving only an auxiliary role. Excessive foreground detail from history frames can introduce attention noise, making it harder for the model to predict the correct action to take on the current screenshot.

In summary, we propose an effective method for partitioning GUI screenshots into foreground and background regions, and reveal a counter-intuitive finding: background information is not dispensable. This insight suggests that when designing efficient token pruning methods for GUI screenshots, one should carefully balance the token budget allocated to foreground and background regions.

\begin{figure*}
    \centering
    \includegraphics[width=0.9\linewidth]{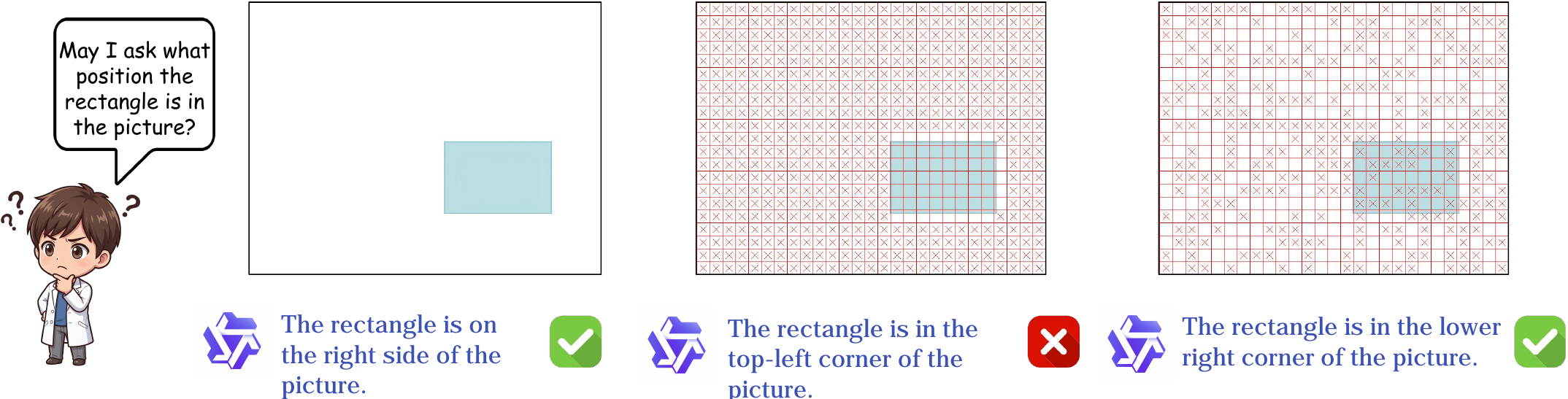}
    \caption{\textbf{A toy example illustrating the effect of token pruning on spatial perception}. Removing background tokens causes the model to lose global spatial references and produce an incorrect answer, whereas random pruning preserves spatial uniformity and maintains correct spatial perception.}
    \label{fig:5}
\end{figure*}

\section{Spatial Perspective: Preserving Spatial Uniformity}

\newcommand{\graytext}[1]{\textcolor[rgb]{.502,.502,.502}{#1}}

\begin{table}[h]
\centering

\resizebox{\columnwidth}{!}{%
\begin{tabular}{lcccc}
\toprule
\textbf{Method} & \textbf{Domain} & \textbf{Task} & \textbf{Website} & \textbf{Average} \\
\midrule
\multicolumn{5}{c}{\textit{\textbf{Qwen2.5-VL-3B}}} \\
\midrule
Upper Bound & 0.4889 & 0.5580 & 0.4697 & 0.5055 \\
\midrule
FastV       & 0.4829 & 0.5483 & 0.4624 & 0.4979 \\
SparseVLM   & 0.4815 & 0.5498 & 0.4573 & 0.4962 \\
DivPrune    & 0.4855 & 0.5430 & 0.4654 & 0.4980 \\
DART        & 0.4776 & 0.5305 & 0.4588 & 0.4890 \\
PDrop       & 0.4826 & 0.5435 & 0.4617 & 0.4959 \\
Random      & 0.4813 & 0.5479 & 0.4668 & 0.4987 \\
\midrule
\multicolumn{5}{c}{\textit{\textbf{Qwen3-VL-2B}}} \\
\midrule
Upper Bound & 0.4863 & 0.5363 & 0.4858 & 0.5028 \\
\midrule
FastV       & 0.4844 & 0.5248 & 0.4836 & 0.4976 \\
SparseVLM   & 0.4819 & 0.5233 & 0.4792 & 0.4948 \\
DivPrune    & 0.4862 & 0.5219 & 0.4821 & 0.4967 \\
DART        & 0.4805 & 0.5185 & 0.4785 & 0.4925 \\
PDrop       & 0.4815 & 0.5233 & 0.4807 & 0.4952 \\
Random      & 0.4860 & 0.5272 & 0.4843 & 0.4992 \\
\bottomrule
\end{tabular}%
}
\caption{Comparison of token pruning methods on Mind2Web ($R=50\%$).}
\label{tab:spatial_mind2web}
\end{table}

\begin{table}[h]
\centering

\resizebox{\columnwidth}{!}{%
\begin{tabular}{lccc}
\toprule
\textbf{Method} & \textbf{type acc} & \textbf{grounding acc} & \textbf{step acc} \\
\midrule
\multicolumn{4}{c}{\textit{\textbf{Qwen2.5-VL-3B}}} \\
\midrule
Upper Bound & 0.8532 & 0.7670 & 0.7113 \\
\midrule
FastV       & 0.8514 & 0.7645 & 0.7084 \\
SparseVLM   & 0.8492 & 0.7602 & 0.7055 \\
DivPrune    & 0.8513 & 0.7657 & 0.7062 \\
DART        & 0.8487 & 0.7614 & 0.7039 \\
PDrop       & 0.8514 & 0.7632 & 0.7044 \\
Random      & 0.8528 & 0.7665 & 0.7115 \\
\midrule
\multicolumn{4}{c}{\textit{\textbf{Qwen3-VL-2B}}} \\
\midrule
Upper Bound & 0.8537 & 0.7562 & 0.7069 \\
\midrule
FastV       & 0.8489 & 0.7583 & 0.7042 \\
SparseVLM   & 0.8496 & 0.7497 & 0.7013 \\
DivPrune    & 0.8522 & 0.7545 & 0.7052 \\
DART        & 0.8529 & 0.7519 & 0.7036 \\
PDrop       & 0.8511 & 0.7540 & 0.7041 \\
Random      & 0.8534 & 0.7580 & 0.7073 \\
\bottomrule
\end{tabular}%
}
\caption{Comparison of token pruning methods on AndroidControl ($R=50\%$).}
\label{tab:spatial_androidcontrol}
\end{table}

In most token pruning methods for MLLMs, pruning is implemented by discarding a subset of visual tokens while preserving the original positional indices of the retained tokens. That is, the remaining tokens are not re-indexed or spatially rearranged after pruning. At first glance, this design appears to preserve spatial information, since each retained token is still associated with its original position. However, we observe that the model's spatial perception can still degrade after token pruning. This raises a natural question: if the positional encodings of the retained tokens remain unchanged, why does the model still lose its ability to reason about the global spatial layout? 

In the Qwen-VL series of multimodal large language models, positional information is modeled via Rotary Position Embedding \cite{RoPE}. For a token at position $m$ with feature vector of dimension $d$, RoPE partitions it into $d/2$ two-dimensional subspaces, where $\theta_k = 10000^{-2k/d}$ is the preset rotation frequency of the $k$-th subspace, and $R(m, \theta_k)$ denotes the 2×2 rotation matrix applied to a token at position $m$ in that subspace:
\begin{equation}
\scalebox{0.85}{$
R(m, \theta_k) = 
\begin{bmatrix} \cos(m\theta_k) & -\sin(m\theta_k) \\ 
\sin(m\theta_k) & \cos(m\theta_k) \end{bmatrix}
$}
\end{equation}
such that:
\begin{equation}
\scalebox{0.85}{$
\begin{bmatrix} v'_{2k} \\ v'_{2k+1} \end{bmatrix} = 
R(m, \theta_k)
\begin{bmatrix} v_{2k} \\ v_{2k+1} \end{bmatrix}
$}
\end{equation}
$R_m$ is then the block-diagonal rotation matrix acting on the full 
$d$-dimensional vector:
\begin{equation}
R_m = \text{diag}(R(m,\theta_0), \ldots, R(m,\theta_{d/2-1}))
\end{equation}
The attention score between tokens at positions $n$ and $m$ is then given by:
\begin{equation}
\scalebox{0.9}{$
\text{AttentionScore}(q_n, k_m) = x_n^T W_q^T R_{m-n} W_k x_m
$}
\end{equation}
where we use the fact that rotation matrices are orthogonal, i.e., $R_n^T = R_{-n}$, and satisfy the angle-addition property, i.e., $R_{-n}R_m = R_{m-n}$. This formula reveals that RoPE computes attention based solely on the relative position difference $m-n$. When the token sequence is complete, the model can indirectly infer the global position of each token through its relative relationships with the other tokens. However, once tokens are pruned, this indirect inference mechanism becomes less reliable. The model can only perceive the position of a retained token through its relationships with other retained tokens, making it difficult to precisely infer its position relative to the global image layout. The more tokens are pruned and the more spatially concentrated the retained tokens are, the more severely the global spatial perception degrades. This analysis extends naturally to M-RoPE and interleaved M-RoPE. A detailed derivation is provided in Appendix~\ref{app:mrope_analysis}.

\begin{figure*}[t]
\centering

\begin{minipage}[t]{0.56\textwidth}
\centering
\vspace{0pt}
\footnotesize
\resizebox{\linewidth}{!}{%
\begin{tabular}{lcccccc}
\toprule
\textbf{Method} & \textbf{General} & \textbf{Single} & \textbf{Webshopping} & \textbf{Install} & \textbf{GoogleApps} & \textbf{Average} \\
\midrule
Upper Bound        
& 0.6496 
& 0.7844 
& 0.6579 
& 0.7366 
& 0.6891 
& 0.7035 \\
\midrule
FastV       & 0.6295 & 0.7964 & 0.6507 & 0.7422 & 0.6792 & 0.6996 \\
SparseVLM   & 0.6235 & 0.7695 & 0.6645 & 0.7334 & 0.7069 & 0.6996 \\
DivPrune    & 0.6259 & 0.7874 & 0.6621 & 0.7318 & 0.6713 & 0.6957 \\
DART        & 0.6259 & 0.7964 & 0.6489 & 0.7310 & 0.6911 & 0.6987 \\
PDrop       & 0.6295 & 0.7844 & 0.6549 & 0.7398 & 0.6733 & 0.6964 \\
Random      & 0.6330 & 0.7934 & 0.6507 & 0.7334 & 0.7030 & 0.7027 \\
\bottomrule
\end{tabular}%
}
\captionof{table}{Comparison of token pruning methods on AITW (Qwen3-VL-2B, $R=50\%$).}
\label{tab:spatial_aitw}
\end{minipage}%
\hfill%
\begin{minipage}[t]{0.40\textwidth}
\centering
\vspace{0pt}
\includegraphics[width=\linewidth]{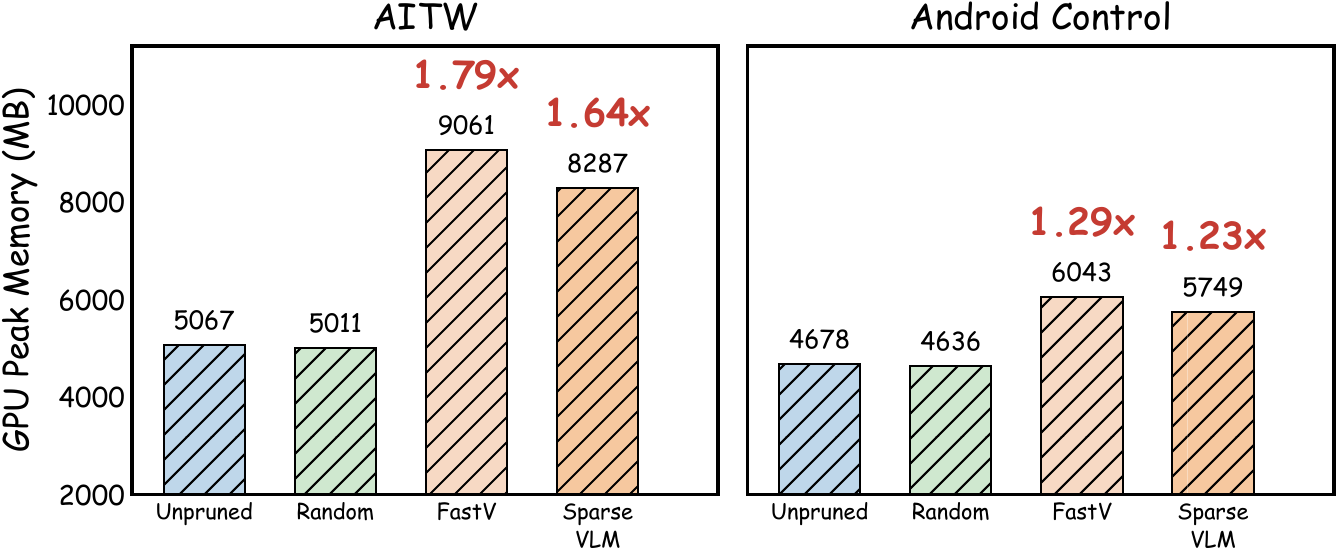}
\captionof{figure}{GPU peak memory comparison of different pruning methods on AITW and AndroidControl (Qwen3-VL-2B, $R=50\%$).}
\label{fig:gpu_mem}
\end{minipage}

\end{figure*}

To concretely illustrate this phenomenon, we construct a toy example shown in Figure~\ref{fig:5}. An image containing a rectangular object on the right side is fed into Qwen with the prompt: "May I ask what position the rectangle is in the picture?" The model correctly answers "on the right side of the image." All background tokens are then removed, retaining only the tokens covering the target object, and the same question is posed. The model's answer changes to "upper left corner." The underlying reason is that the pairwise relative positional relationships among the tokens inside the rectangle remain intact, so the model can still recognize the shape. However, the removal of background tokens eliminates the spatial references needed to determine the rectangle's global position, leading to an erroneous location judgment. In GUI navigation, where interface element positions directly determine the agent's interaction targets, such degradation can directly impair task execution.

By design, random pruning maintains an approximately uniform spatial distribution: the removed tokens are scattered across the entire image, and so are the retained tokens. The retained tokens thus continue to serve as \textit{spatial anchors} distributed across the global layout, providing the model with sufficient positional references and better preserving the spatial correspondence between tokens and the global context. In contrast, most carefully designed pruning methods do not explicitly optimize for spatial uniformity, which may impair the model's global spatial perception.

As shown in Tables~\ref{tab:spatial_mind2web},~\ref{tab:spatial_androidcontrol}, and~\ref{tab:spatial_aitw}, under the same token budget constraint, random pruning achieves competitive performance. This further confirms that maintaining spatial uniformity is an important consideration in token pruning design for GUI agent navigation. Results under more aggressive retention budgets are provided in Appendix~\ref{app:spatial_uniformity_validation}.

In addition to preserving spatial uniformity, random pruning slightly reduces GPU peak memory, whereas attention-based pruning methods introduce substantial additional memory overhead.
Attention-based methods such as FastV and SparseVLM require explicit access to attention maps, making them incompatible with efficient attention operators such as FlashAttention. While such overhead may be less noticeable in single-image QA scenarios, it becomes substantially amplified in GUI agent navigation, where multiple historical screenshots are included in the input. As shown in Figure~\ref{fig:gpu_mem}, when retaining 50\% of the visual tokens for each historical screenshot, FastV and SparseVLM increase GPU peak memory from 5067 MB to 9061 MB and 8287 MB on AITW, and from 4678 MB to 6043 MB and 5749 MB on AndroidControl, respectively. In comparison, random pruning does not require computing attention maps, and therefore slightly reduces GPU peak memory.

\section{Conclusion}
In this paper, we study token pruning for GUI agent navigation through two questions: \textit{where to prune} and \textit{how to prune}. For the first question, we identify a system-level redundancy in GUI agent navigation: the same screenshot is often fed into the model across multiple steps, so its ViT features can be cached and reused. This cache-and-reuse mechanism is not tied to any specific pruning strategy. Rather, it serves as a broadly applicable acceleration technique for GUI agents that use historical screenshots. In contrast, reducing tokens before ViT encoding causes performance degradation, suggesting that token pruning should mainly be applied in the subsequent LLM. For the second question, we show that effective LLM-side pruning should preserve both semantic balance and spatial structure: background tokens should not be blindly discarded, and retained tokens should remain spatially uniform. We hope these findings provide useful guidance for efficient GUI agent design.

\bibliography{aaai2027}

\clearpage
\appendix
\setcounter{equation}{0}

\setcounter{secnumdepth}{2}

\renewcommand{\thesection}{\Alph{section}}
\renewcommand{\thesubsection}{\thesection.\arabic{subsection}}
\renewcommand{\thesubsubsection}{\thesubsection.\arabic{subsubsection}}

\section*{Appendix Overview}
This appendix provides additional details and analyses to supplement the discussion in the main paper. Appendix~\ref{app:dataset_details} describes the datasets used in our experiments. Appendix~\ref{app:training_eval_details} reports the training setup, hardware, FLOPs and memory measurement, and evaluation metrics. Appendix~\ref{app:segmentation_visualization} visualizes the edge-based foreground--background partition. Appendix~\ref{app:pruning_layer} studies the choice of pruning layer. Appendix~\ref{app:mrope_analysis} extends the RoPE analysis in the main paper to M-RoPE and interleaved M-RoPE. Appendix~\ref{app:spatial_uniformity_validation} further validates spatial uniformity under a more aggressive pruning budget. Appendix~\ref{app:comparison_methods} summarizes the compared token pruning methods. Appendix~\ref{app:kb_dagger} details the construction of KB$^\dagger$. Appendix~\ref{app:llm-kv-cache-analysis} analyzes why LLM KV-cache reuse is more restrictive than ViT feature caching. Appendix~\ref{app:current_screenshot_pruning} analyzes the effect of pruning the current screenshot. Appendix~\ref{app:cache_pruning}  demonstrates the effectiveness of jointly applying ViT feature caching and LLM-side token pruning.

\section{Dataset Details}
\label{app:dataset_details}
\paragraph{AITW.}
Android in the Wild (AITW) is a large-scale Android device control benchmark collected from real human demonstrations. It consists of 715k episodes spanning 30k unique task instructions. The dataset is organized into five subsets: \textit{General}, \textit{GoogleApps}, \textit{Install}, \textit{Webshopping}, and \textit{Single}. For a fair evaluation on the AITW dataset, we adopt the instruction-wise split protocol introduced by SeeClick \cite{seeclick}. We also follow the same data preprocessing settings as SeeClick.

\paragraph{Mind2Web.}
Mind2Web is a real-world web navigation benchmark comprising over 2,000 open-ended tasks collected from 137 websites spanning 31 domains. Each task is paired with a natural language instruction and a sequence of crowdsourced actions. The benchmark is evaluated under three generalization settings: cross-task (M-t), cross-website (M-w), and cross-domain (M-d), making it a testbed for assessing the generalizability of web agents across scenarios.

\paragraph{AndroidControl.}
AndroidControl is a large-scale and diverse Android UI control benchmark consisting of 15,283 human demonstrations across 833 Android apps. A distinctive feature of this benchmark is that each task instance includes both high-level and low-level human-generated instructions, enabling evaluation at different levels of task abstraction. It is currently the most diverse UI control dataset in terms of app coverage, containing 14,548 unique tasks. 

\paragraph{GUI-Odyssey.}
GUI-Odyssey is a comprehensive benchmark specifically designed for cross-app navigation on mobile devices, a scenario that has been largely overlooked by prior datasets. It consists of 8,334 episodes with an average of 15.3 steps per episode, covering 6 mobile devices, 212 distinct apps, and 1,357 app combinations across 6 types of cross-app tasks. The longer episode length compared to AITW and AndroidControl makes it suited for evaluating agents in long-horizon sequential decision-making scenarios where historical context plays a critical role.

\section{Training and Evaluation Details}
\label{app:training_eval_details}
\subsection{Training Details}
Since the base models are not natively trained for GUI agent tasks, we first fine-tune both Qwen2.5-VL-3B and Qwen3-VL-2B on all four benchmarks using supervised fine-tuning (SFT) to establish strong task-specific baselines. We fine-tune all models using LoRA adaptation.
Across all four benchmarks, we adopt a unified training configuration. Specifically, the LoRA rank is set to $r=8$, with scaling factor $\alpha=16$ and dropout rate 0. LoRA is applied to all linear layers of the LLM, while excluding the token embedding layer and the output head. The vision encoder is kept frozen, and the multimodal merger module is not updated. Only the LoRA adapter parameters in the LLM are optimized. Training is conducted with DeepSpeed ZeRO-2 optimization. We enable bfloat16 mixed-precision training while disabling FP16.
FlashAttention-2 remains enabled during training. The per-device batch size varies by dataset:
8 for AITW, 2 for Mind2Web and AndroidControl,
and 4 for GUI-Odyssey. The learning rate is set to $3\times10^{-5}$ for AITW and GUI-Odyssey,
$5\times10^{-4}$ for Mind2Web, and $3\times10^{-4}$ for AndroidControl. For each benchmark, we independently fine-tune the two base models using only the benchmark's training split and evaluate the resulting checkpoints on the corresponding official evaluation split. No data are mixed across benchmarks. Token pruning is applied exclusively at inference time without additional retraining.

\begin{figure*}[t]
    \centering
    \includegraphics[width=1.0\linewidth]{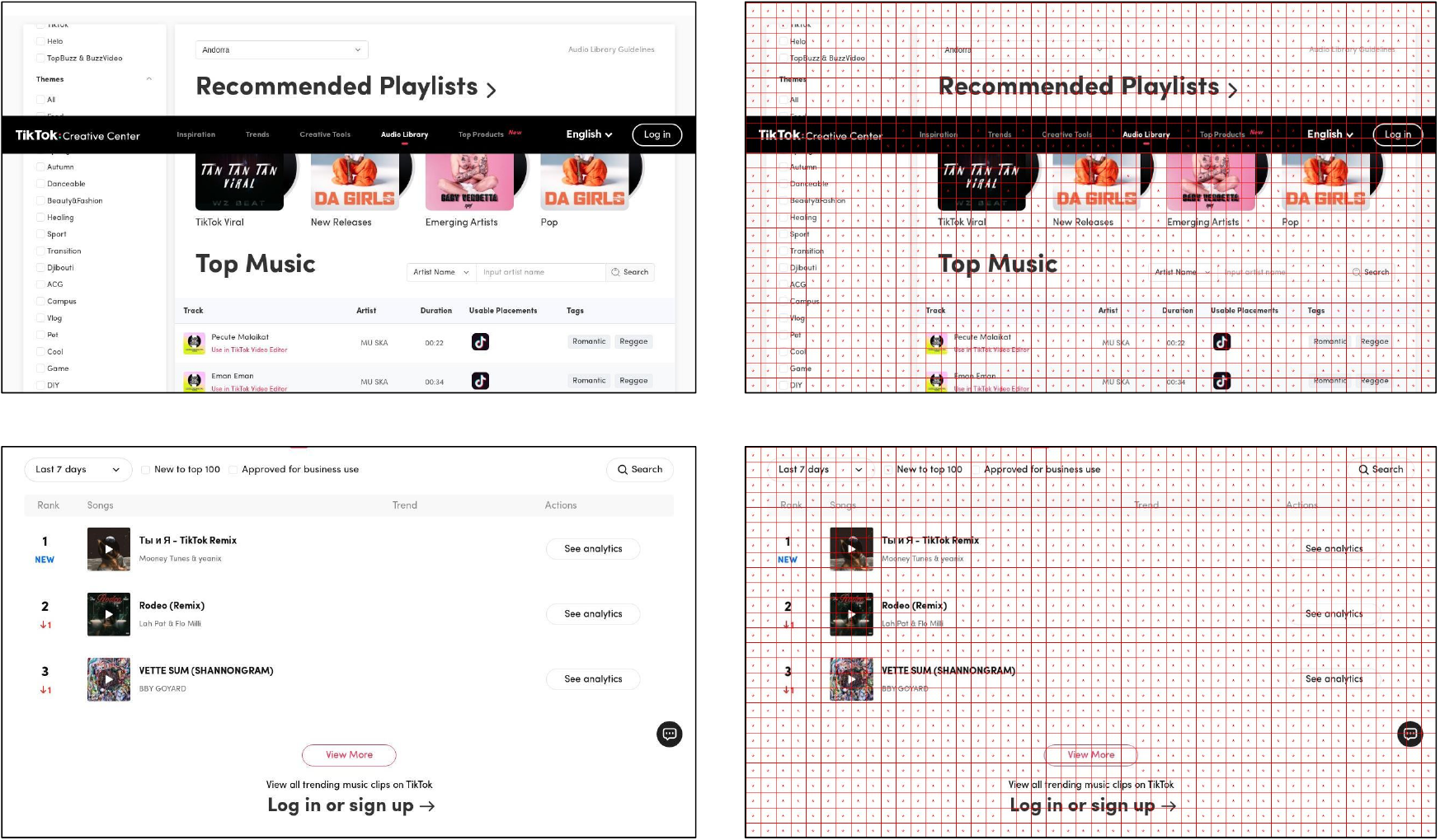}
    \caption{Visualization of edge-based foreground--background segmentation on Mind2Web screenshots. The same parameter settings
($\tau_e = 50$, $\tau_r = 0.01$) as in the main paper produce accurate
results across diverse web interface layouts.}
    \label{fig:edge}
\end{figure*}

\subsection{Hardware}
Model training is conducted on four NVIDIA H200 GPUs. Evaluations are performed on an NVIDIA RTX-3090 for Qwen3-VL-2B and an NVIDIA A100 for Qwen2.5-VL-3B.

\subsection{FLOPs Measurement}
We report FLOPs measured using \texttt{torch.profiler}, 
including ViT encoding, LLM prefill, and decoding.

\subsection{GPU Peak Memory Measurement}
We report GPU peak memory usage measured via 
\texttt{torch.cuda.max\_memory\_allocated()}.

\subsection{Evaluation Metrics}
Following SimpAgent, we adopt the step success rate as the evaluation metric across all benchmarks. For AndroidControl, we additionally report type and grounding accuracy.

\section{Visualization of Segmentation}
\label{app:segmentation_visualization}

Figure~\ref{fig:edge} presents the Sobel edge detection results on screenshots from the Mind2Web dataset, complementing the AITW example shown in the main paper. As can be observed, the edge-based segmentation method with the same parameter settings ($\tau_e = 50$, $\tau_r = 0.01$) produces clean and accurate foreground-background partitions across visually diverse web interface screenshots. These results demonstrate that the proposed approach generalizes well across datasets.

\section{Choice of Pruning Layer}
\label{app:pruning_layer}
\begin{figure*}
    \centering
    \includegraphics[width=1\linewidth]{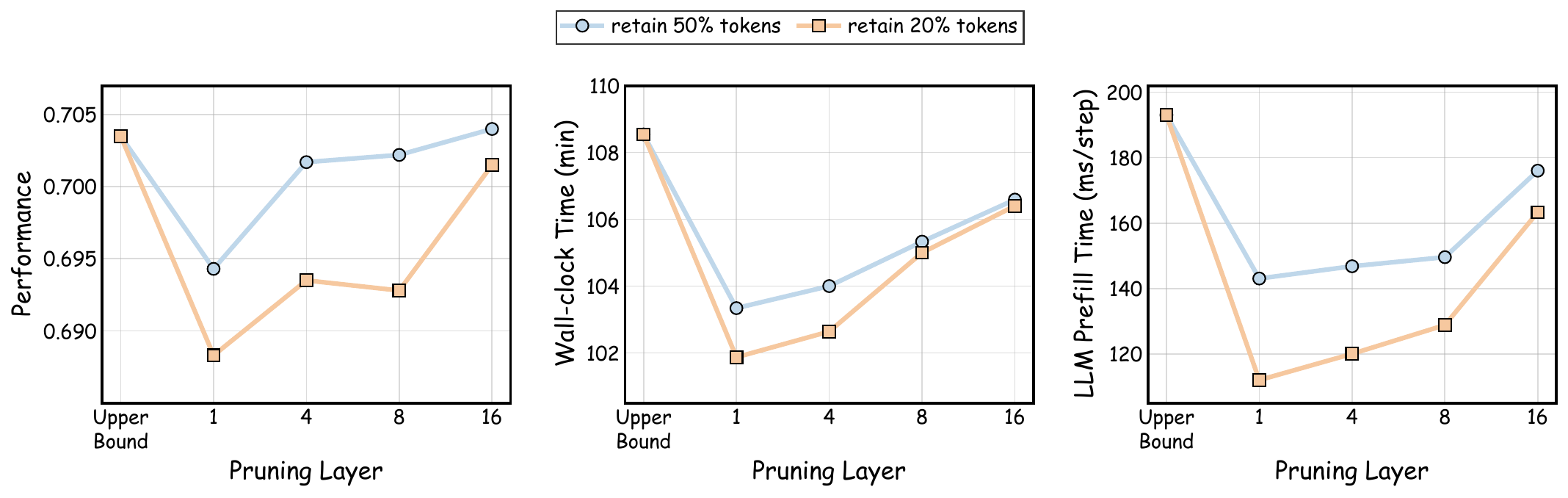}
    \caption{Effect of pruning layer selection under different retention ratios. Upper Bound denotes the setting without pruning.}
    \label{fig:layer}
\end{figure*}

\begin{table*}[t]
\centering
\scriptsize
\resizebox{0.75\textwidth}{!}{
\begin{tabular}{lcccccc}
\toprule
Method & General & Single & Webshopping & Install & GoogleApps & Average \\
\midrule
Upper Bound      & 0.6496 & 0.7844 & 0.6579 & 0.7366 & 0.6891 & 0.7035 \\

\midrule
\multicolumn{7}{c}{\textit{30\% Token Retention}} \\
\midrule
FastV      & 0.6211 & 0.7784 & 0.6591 & 0.7294 & 0.6911 & 0.6958 \\
SparseVLM  & 0.6211 & 0.7575 & 0.6603 & 0.7287 & 0.6931 & 0.6921 \\
DivPrune   & 0.6211 & 0.7665 & 0.6591 & 0.7294 & 0.6812 & 0.6915 \\
DART       & 0.6140 & 0.7725 & 0.6292 & 0.7255 & 0.6871 & 0.6857 \\
PDrop      & 0.6283 & 0.7575 & 0.6573 & 0.7271 & 0.6970 & 0.6934 \\
Random     & 0.6152 & 0.7874 & 0.6567 & 0.7390 & 0.6911 & 0.6979 \\

\midrule
\multicolumn{7}{c}{\textit{10\% Token Retention}} \\
\midrule
FastV      & 0.6045 & 0.7635 & 0.6382 & 0.7207 & 0.6851 & 0.6824 \\
SparseVLM  & 0.6033 & 0.7485 & 0.6352 & 0.7103 & 0.6832 & 0.6761 \\
DivPrune   & 0.6093 & 0.7515 & 0.6459 & 0.7215 & 0.6851 & 0.6827 \\
DART       & 0.5938 & 0.7545 & 0.6220 & 0.6999 & 0.6634 & 0.6667 \\
PDrop      & 0.6188 & 0.7545 & 0.6286 & 0.7175 & 0.6812 & 0.6801 \\
Random     & 0.6140 & 0.7605 & 0.6274 & 0.7223 & 0.6970 & 0.6842 \\
\bottomrule
\end{tabular}}
\caption{Comparison of token pruning methods on AITW under 10\% and 30\% token-retention budgets. Upper Bound denotes the model without token pruning.}
\label{tab:spatial_aitw_different_budgets}
\end{table*}

We further study how the layer at which token pruning is applied affects the trade-off between performance and inference efficiency. The results show that the pruning layer has a clear impact on both accuracy preservation and acceleration.

When pruning is applied too early, such as at the 1st layer, the model obtains slightly lower inference cost but suffers from a more noticeable performance drop. In contrast, pruning at the 4th layer consistently achieves better performance under both token retention ratios, while introducing only a small increase in wall-clock time and LLM prefill time compared with pruning at the 1st layer. This suggests that pruning too early is suboptimal. In shallow layers, attention is still relatively dispersed, and token information has not been sufficiently aggregated into representative anchor tokens. 

On the other hand, when pruning is applied at deeper layers, such as the 8th or 16th layer, the model performance generally improves, but the acceleration benefit becomes much smaller. This is because a large portion of the LLM prefill computation has already been performed before pruning, limiting the overall speedup that can be achieved.

Considering both performance preservation and inference efficiency, we choose the 4th layer as the default pruning layer. When retaining 50\% of the tokens, pruning at the 4th layer preserves about 99.7\% of the Upper Bound performance, while reducing LLM prefill time by approximately 24.0\% and wall-clock time by approximately 4.2\%. When retaining 20\% of the tokens, it still preserves about 98.6\% of the Upper Bound performance, while reducing LLM prefill time by approximately 37.8\% and wall-clock time by approximately 5.4\%. Detailed metrics are shown in Figure~\ref{fig:layer}. Therefore, pruning at the 4th layer provides a good balance between maintaining model performance and improving inference efficiency.

\section{Extension of the RoPE Analysis to M-RoPE and Interleaved M-RoPE}
\label{app:mrope_analysis}

The main paper uses one-dimensional RoPE to explain how token pruning affects spatial perception. In this section, we extend the derivation to Multimodal Rotary Position Embedding (M-RoPE), which decomposes positional encoding into temporal, height, and width components.

For a token $m$ at position
\begin{equation}
\mathbf{p}_m = (t_m, h_m, w_m),
\end{equation}
$t_m$, $h_m$, and $w_m$ denote its temporal, height, and width position indices, respectively. M-RoPE assigns each of the $d/2$ two-dimensional rotary subspaces to one of these three positional axes. Let $a(k)\in\{t,h,w\}$ denote the axis assigned to the $k$-th subspace, and define
\begin{equation}
\xi_{m,k} =
\begin{cases}
t_m, & a(k)=t,\\
h_m, & a(k)=h,\\
w_m, & a(k)=w.
\end{cases}
\end{equation}

Following the notation used in the main paper, the rotation matrix for the $k$-th two-dimensional subspace is
\begin{equation}
R(\xi_{m,k},\theta_k)
=
\begin{bmatrix}
\cos(\xi_{m,k}\theta_k) & -\sin(\xi_{m,k}\theta_k)\\
\sin(\xi_{m,k}\theta_k) & \cos(\xi_{m,k}\theta_k)
\end{bmatrix}.
\end{equation}
The M-RoPE rotation matrix acting on the complete $d$-dimensional vector can therefore be written as
\begin{equation}
\mathcal{R}_{\mathbf{p}_m}
=
\operatorname{diag}
\left(
R(\xi_{m,0},\theta_0),
\ldots,
R(\xi_{m,d/2-1},\theta_{d/2-1})
\right).
\end{equation}

In standard M-RoPE, the rotary subspaces associated with the temporal, height, and width axes are grouped together. Interleaved M-RoPE instead distributes these subspaces across different frequency bands. The two variants differ only in the axis assignment $a(k)$, and the following relative-position derivation applies to both.

For two tokens at positions $\mathbf{p}_n$ and $\mathbf{p}_m$, their queries and keys after applying M-RoPE are
\begin{equation}
\widetilde{q}_n
=
\mathcal{R}_{\mathbf{p}_n}W_qx_n,
\qquad
\widetilde{k}_m
=
\mathcal{R}_{\mathbf{p}_m}W_kx_m.
\end{equation}
Their attention score is therefore
\begin{equation}
\operatorname{AttentionScore}(q_n,k_m)
=
x_n^T W_q^T
\mathcal{R}_{\mathbf{p}_n}^T
\mathcal{R}_{\mathbf{p}_m}
W_k x_m.
\end{equation}

For each two-dimensional rotary subspace, we have
\begin{equation}
R(\xi_{n,k},\theta_k)^T
R(\xi_{m,k},\theta_k)
=
R(\xi_{m,k}-\xi_{n,k},\theta_k).
\end{equation}
Consequently,
\begin{equation}
\mathcal{R}_{\mathbf{p}_n}^T
\mathcal{R}_{\mathbf{p}_m}
=
\mathcal{R}_{\mathbf{p}_m-\mathbf{p}_n},
\end{equation}
where
\begin{equation}
\mathbf{p}_m-\mathbf{p}_n
=
(t_m-t_n,\;h_m-h_n,\;w_m-w_n).
\end{equation}
The attention score can thus be written as
\begin{equation}
\operatorname{AttentionScore}(q_n,k_m)
=
x_n^T W_q^T
\mathcal{R}_{\mathbf{p}_m-\mathbf{p}_n}
W_k x_m.
\end{equation}

This result shows that, similar to one-dimensional RoPE, attention under M-RoPE depends on the relative positional differences between tokens, but the one-dimensional difference $m-n$ is extended to the temporal, height, and width dimensions. For visual tokens within the same screenshot, the temporal position remains constant, such that $t_m-t_n=0$. Their positional relationships are therefore primarily determined by the height difference $h_m-h_n$ and the width difference $w_m-w_n$.

\section{Further Validation of Spatial Uniformity}
\label{app:spatial_uniformity_validation}

We further evaluate different token pruning methods on the AITW dataset using Qwen3-VL-2B under two aggressive token-retention budgets, retaining only 30\% or 10\% of the visual tokens in each screenshot. As shown in Table~\ref{tab:spatial_aitw_different_budgets}, random pruning achieves the highest average score among all pruning methods under both settings, reaching 0.6979 at a 30\% retention ratio and 0.6842 at a 10\% retention ratio. Notably, its performance remains close to that of the unpruned model even under the 30\% budget. These consistent results further support our observation in the main text that preserving the spatial uniformity of retained tokens is an important consideration when designing token pruning methods for GUI agent navigation.

\section{Comparison Methods}
\label{app:comparison_methods}
We compare against five representative token pruning methods, including FastV, SparseVLM, PDrop, DART, and DivPrune, which cover attention-based, text-guided, redundancy-based, and diversity-based pruning strategies.

\textbf{FastV} is a plug-and-play, training-free method that identifies inefficient attention phenomena in LVLMs, observing that visual tokens 
receive significantly less attention than text tokens in deep layers. It learns adaptive attention patterns in early layers and prunes a fixed ratio of visual tokens with the lowest average attention scores in subsequent layers, achieving substantial 
FLOPs reduction without retraining.

\textbf{SparseVLM} is a text-guided, training-free token sparsification framework. It first selects visual-relevant text tokens as "raters" via cross-modal attention from the decoder layers, then uses the attention from 
these selected text tokens to measure the significance of visual tokens and 
progressively prunes the less important ones. A rank-based strategy is further introduced to adaptively determine sparsification ratios, and a token recycling mechanism is employed to compress pruned information into retained tokens.

\textbf{PDrop} is motivated by the empirical observation that visual token redundancy progressively increases in deeper 
layers of LVLMs. It partitions the model into several stages and drops a predefined 
ratio of image tokens at the end of each stage based on attention scores, forming a 
pyramid-like token structure across layers. This strategy can be applied as a plug-and-play inference accelerator or integrated into training to reduce both inference FLOPs and training time.

\textbf{DART} challenges 
the conventional wisdom of importance-based pruning, showing that selecting tokens based on importance metrics can lead to inferior performance compared to random pruning and is incompatible with efficient attention operators. Instead, DART prunes tokens based on their duplication with other tokens: it selects a small subset of pivot tokens (no more than 2\% of total tokens) and retains only those vision tokens with low cosine similarity to the pivots, ensuring minimal information loss while maintaining compatibility with FlashAttention.

\textbf{DivPrune} formulates visual token pruning as a Max-Min Diversity Problem, where the goal is to select a subset of tokens such that the minimum pairwise distance among selected tokens is maximized. This diversity-based approach reduces redundancy among retained tokens without requiring fine-tuning, and achieves state-of-the-art accuracy while reducing both latency and GPU memory usage.

\textbf{Random} uniformly samples visual tokens from each historical
screenshot according to the target retention ratio, without relying on
attention scores, token importance, or feature redundancy. For each
experimental setting, we conduct three independent runs without fixing
the random seed and report the run with the median overall score, rather
than averaging the results across the three runs.

For a fair comparison, all methods prune historical visual tokens only at the fourth LLM layer. SparseVLM and PDrop, which originally perform pruning across multiple layers, are therefore adapted to this single-layer setting while retaining their respective mechanisms. 

\section{Construction of KB$^\dagger$}
\label{app:kb_dagger}

In the semantic analysis of foreground and background regions, we introduce KB$^\dagger$ as a budget-matched background-only control. The motivation of KB$^\dagger$ is twofold. First, since background patches usually account for a larger proportion than foreground patches, directly comparing KB with KF may be unfair due to the larger token budget of KB. Second, background patches are often more spatially dispersed than foreground patches. If background patches are randomly subsampled, the retained background tokens may still cover a broad spatial area, making it difficult to determine whether the advantage of background retention comes from background information itself or from its broader spatial coverage. Therefore, KB$^\dagger$ is designed to match the token budget of KF while making the retained background patches as spatially concentrated as possible.

Specifically, for each historical screenshot, we first apply the same edge-based foreground-background partitioning procedure, obtaining a foreground patch set $\mathcal{F}_j$ and a background patch set $\mathcal{B}_j$ for the $j$-th historical screenshot. Let $K_j = |\mathcal{F}_j|$ denote the number of foreground patches retained by KF for this screenshot. Given the visual-token grid of size $H \times W$, each background patch $p_i \in \mathcal{B}_j$ has a grid coordinate $(r_i, c_i)$. We define the geometric center of the visual-token grid as
\begin{equation}
(r_0, c_0) = \left(\frac{H-1}{2}, \frac{W-1}{2}\right).
\end{equation}
For each background patch, we compute its Euclidean distance to the grid center:
\begin{equation}
d_i = \sqrt{(r_i-r_0)^2 + (c_i-c_0)^2}.
\end{equation}
We then sort all background patches in ascending order according to $d_i$ and select background patches from the center outward.

In most cases, if $|\mathcal{B}_j| \ge K_j$, KB$^\dagger$ retains the nearest $K_j$ background patches:
\begin{equation}
\mathcal{B}^{\dagger}_j
=
\operatorname{TopK}_{p_i \in \mathcal{B}_j}(-d_i, K_j).
\end{equation}
Equivalently, this procedure starts from the center of the visual-token grid and gradually expands outward until the number of retained background patches matches the token budget of KF for this screenshot.

However, for some screenshots, the number of foreground patches may exceed 50\% of all visual patches. In such cases, the number of background patches is smaller than the foreground budget, i.e., $|\mathcal{B}_j| < K_j$. Even retaining all background patches from this screenshot cannot match the KF budget. For these cases, we retain all background patches in $\mathcal{B}_j$ and compute the budget deficit:
\begin{equation}
\Delta_j = K_j - |\mathcal{B}_j|.
\end{equation}
The missing budget is then reassigned to temporally adjacent historical screenshots within the same inference step. We prioritize nearby historical screenshots that still have unselected background patches. For a recipient screenshot, the additional budget is filled by continuing the same center-outward selection order, i.e., by selecting the next nearest background patches that have not yet been retained.

With this redistribution, KB$^\dagger$ matches the token budget of KF at the inference-step level:
\begin{equation}
\sum_j |\mathcal{B}^{\dagger}_j|
=
\sum_j |\mathcal{F}_j|.
\end{equation}
Therefore, although KB$^\dagger$ may not strictly match the KF budget for every individual screenshot when a screenshot lacks sufficient background patches, it ensures that the total number of retained historical visual tokens is the same as KF for each inference step.

This construction removes both the token-budget advantage and the spatial-uniformity advantage that background patches may naturally have over foreground patches. Therefore, KB$^\dagger$ provides a stricter control for examining whether background regions remain useful beyond these confounding factors, supporting the view that they encode useful historical information such as environmental context and interface state transitions.

\begin{table*}[t]
\centering
\small
\resizebox{\textwidth}{!}{
\begin{tabular}{lccccccc}
\toprule
Method
& General
& Single
& Webshopping
& Install
& GoogleApps
& Avg.
& Wall-clock Time (min) \\
\midrule

No Cache
& 0.6496
& 0.7844
& 0.6579
& 0.7366
& 0.6891
& 0.7035
& 108.54 \\

ViT Feature Cache
& 0.6390
& 0.7784
& 0.6615
& 0.7310
& 0.7050
& 0.7030
& 95.70 \\

ViT + Strict LLM KV Cache
& 0.6532
& 0.7814
& 0.6579
& 0.7310
& 0.6832
& 0.7013
& 92.89 \\

ViT + Segment-level LLM KV Cache
& 0.5190
& 0.7665
& 0.3708
& 0.5196
& 0.5406
& 0.5433
& 105.48 \\

\bottomrule
\end{tabular}
}
\caption{\textbf{Ablation on KV cache reuse on AITW.}
ViT feature caching and strict LLM KV caching largely preserve task performance while reducing wall-clock inference time. In contrast, segment-level LLM KV reuse substantially degrades performance because the reused states do not correspond to exact shared prefixes.}
\label{tab:llm-kv-cache-ablation-aitw}
\end{table*}

\begin{table*}[t]
\centering
\small
\begin{tabular}{c l l}
\toprule
Step & Prompt structure with $H=2$ & Safe LLM KV reuse from previous step \\
\midrule
$t=0$ 
& $[U, O_0]$ 
& -- \\

$t=1$ 
& $[U, O_0, A_0, O_1]$ 
& $[U, O_0]$ \\

$t=2$ 
& $[U, O_0, A_0, O_1, A_1, O_2]$ 
& $[U, O_0, A_0, O_1]$ \\

$t=3$ 
& $[U, O_1, A_1, O_2, A_2, O_3]$ 
& only the initial text prefix before $O_1$ \\

$t=4$ 
& $[U, O_2, A_2, O_3, A_3, O_4]$ 
& only the initial text prefix before $O_2$ \\
\bottomrule
\end{tabular}
\caption{Illustration of strict LLM KV cache reuse with a 2-history window. Before the window slides, the prompt grows by appending new action and screenshot segments, so the previous prompt is an exact prefix of the current prompt. After the window slides at $t=3$, the oldest pair $(O_0,A_0)$ is removed, and the remaining screenshots appear under different causal prefixes. Therefore, their previous KV states are no longer valid.}
\label{tab:kv-cache-window-example}
\end{table*}

\section{Analysis of LLM KV Cache Reuse}
\label{app:llm-kv-cache-analysis}

Our main caching strategy reuses the visual features produced by the vision encoder. This reuse is safe because the ViT processes each screenshot independently: after the same preprocessing pipeline, the visual representation is determined by the screenshot itself and does not depend on the timestep or surrounding textual context.

A natural question is whether we can further reuse the key-value (KV) caches inside the LLM. We analyze this possibility and find that LLM KV states are fundamentally different from ViT features. ViT features are screenshot-level representations, whereas LLM KV states are prefix-dependent representations tied to the exact sequence in which the corresponding tokens appear.

For a causal Transformer layer, the key and value states of token $i$ at layer $\ell$ can be written as
\begin{equation}
    K_i^{(\ell)} = W_K^{(\ell)} h_i^{(\ell)}, \qquad
    V_i^{(\ell)} = W_V^{(\ell)} h_i^{(\ell)},
\end{equation}
where $h_i^{(\ell)}$ is the hidden state before the self-attention module at layer $\ell$. The KV states of a screenshot token or an action token depend not only on its own, but also on its exact causal prefix.

This property makes the conditions for LLM KV reuse more restrictive than those for ViT feature reuse. Let $U$ denote the user instruction, $O_t$ the screenshot at step $t$, and $A_t$ the action taken after observing $O_t$. With a history window of size $H$, the input prompt at step $t$ can be abstractly written as
\begin{equation}
\begin{aligned}
P_t = [&\, U,\,
O_s, A_s,\,
O_{s+1}, A_{s+1},\,
\ldots, \\
& O_{t-1}, A_{t-1},\,
O_t \,].
\end{aligned}
\end{equation}
where $s=\max(0,t-H)$.
LLM KV reuse is safe only when the reused states correspond to an exact shared prefix between prompts. That is, a cached KV state can be reused only if the current prompt contains exactly the same preceding token sequence, visual embeddings, position indices, and attention mask.

\paragraph{Example with a 2-history window.}
In our Mind2Web setting, the model keeps up to two historical screenshots, i.e., $H=2$. The first few steps of an episode are append-only: each new prompt extends the previous one by appending the newly executed action and the next screenshot. In this stage, strict prefix KV reuse is valid because the previous prompt is an exact prefix of the current prompt. However, once the history window is full, the next step removes the oldest screenshot-action pair. The prompt is no longer an extension of the previous prompt; instead, the entire history window shifts.

This example highlights the key difference between ViT feature reuse and LLM KV reuse. The screenshot $O_1$ appears in both $P_2$ and $P_3$. For ViT caching, this is harmless: the visual feature of $O_1$ is tied to the screenshot itself. For LLM KV caching, however, the KV states of the tokens corresponding to $O_1$ are not the same in the two prompts. In $P_2$, $O_1$ is preceded by $[U,O_0,A_0]$ and appears as the second screenshot in the prompt. In $P_3$, $O_1$ becomes the first historical screenshot and is preceded only by the initial instruction prefix. Thus, reusing the old KV states of $O_1$ would inject hidden states computed under a different causal context.

\begin{table}
\centering
\small
\resizebox{\columnwidth}{!}{
\begin{tabular}{lcccc}
\toprule
Method
& Domain
& Task
& Website
& Avg. \\
\midrule

No Cache
& 0.4863
& 0.5363
& 0.4858
& 0.5028 \\

ViT Feature Cache
& 0.4884
& 0.5363
& 0.4829
& 0.5025 \\

ViT + Strict LLM KV Cache
& 0.4918
& 0.5368
& 0.4902
& 0.5063 \\

ViT + Segment-level LLM KV Cache
& 0.2584
& 0.2362
& 0.2210
& 0.2385 \\

\bottomrule
\end{tabular}
}
\caption{\textbf{Ablation on KV cache reuse on Mind2Web.}
ViT feature caching and strict LLM KV caching preserve performance, whereas segment-level LLM KV reuse causes a substantial performance drop.}
\label{tab:llm-kv-cache-ablation-mind2web}
\end{table}

\begin{table*}[t]
\centering
\scriptsize
\resizebox{0.75\textwidth}{!}{
\begin{tabular}{lcccccc}
\toprule
Method & General & Single & Webshopping & Install & GoogleApps & Average \\
\midrule
Upper Bound
& 0.6496
& 0.7844
& 0.6579
& 0.7366
& 0.6891
& 0.7035 \\

\midrule
FastV-0.5
& 0.6200
& 0.7246
& 0.6226
& 0.7031
& 0.6356
& 0.6612 \\

FastV-0.4
& 0.5903
& 0.7216
& 0.5927
& 0.6680
& 0.6178
& 0.6381 \\

FastV-0.3
& 0.5606
& 0.7066
& 0.5694
& 0.6097
& 0.5941
& 0.6081 \\

FastV-0.2
& 0.5131
& 0.6407
& 0.4928
& 0.5435
& 0.5406
& 0.5461 \\
\bottomrule
\end{tabular}}
\caption{Performance of FastV when applied to the current screenshot
on AITW under different visual-token retention ratios. The suffix
denotes the token-retention ratio. Upper Bound denotes the model
without token pruning.}
\label{tab:current_screenshot_pruning}
\end{table*}

To validate this analysis, we compare different caching strategies on AITW and Mind2Web. The first strategy is ViT feature caching, which reuses screenshot-level vision features and is theoretically safe. The second strategy is strict LLM KV cache reuse, where KV states are reused only for exact shared prefixes. This setting is theoretically safe, but it can reuse only a limited portion of the prompt, mainly before the history window starts sliding. The third strategy is segment-level LLM KV cache reuse, where KV states of historical screenshot and action segments are forcibly reused even after the history window slides. This strategy reuses more internal states, but it is not theoretically valid because the reused KV states no longer correspond to the current causal context.

As shown in Tables~\ref{tab:llm-kv-cache-ablation-aitw}
and~\ref{tab:llm-kv-cache-ablation-mind2web}, ViT feature caching preserves performance on both AITW and Mind2Web. Strict LLM KV cache reuse also maintains comparable performance, consistent with its theoretical validity under exact-prefix reuse. On AITW, combining ViT feature caching with strict LLM KV cache reuse reduces the wall-clock time from 108.54 to 92.89 minutes. However, the additional speedup over ViT feature caching alone remains modest, because exact KV reuse is applicable only to the first few steps of each episode, where the input sequences share identical prefixes. In contrast, segment-level LLM KV cache reuse leads to a substantial performance drop, decreasing the average score from 0.7035 to 0.5433 on AITW and from 0.5028 to 0.2385 on Mind2Web. Moreover, such segment-level reuse requires 105.48 minutes of wall-clock time. This is because the additional overhead of storing, retrieving, and managing segment-level KV states largely offsets the computational savings achieved through cache reuse.

This analysis explains why our final method caches ViT outputs rather than aggressively reusing internal LLM KV states across historical GUI segments. Vision features provide reusable screenshot-level representations, while LLM KV states encode both token content and the exact causal prefix. As a result, LLM KV caching is safely applicable only to exact shared prefixes and cannot be generally applied to historical GUI segments after the sliding window changes.

\section{Analysis of Current Screenshot Pruning}
\label{app:current_screenshot_pruning}

\begin{figure}
    \centering
    \includegraphics[width=1\linewidth]{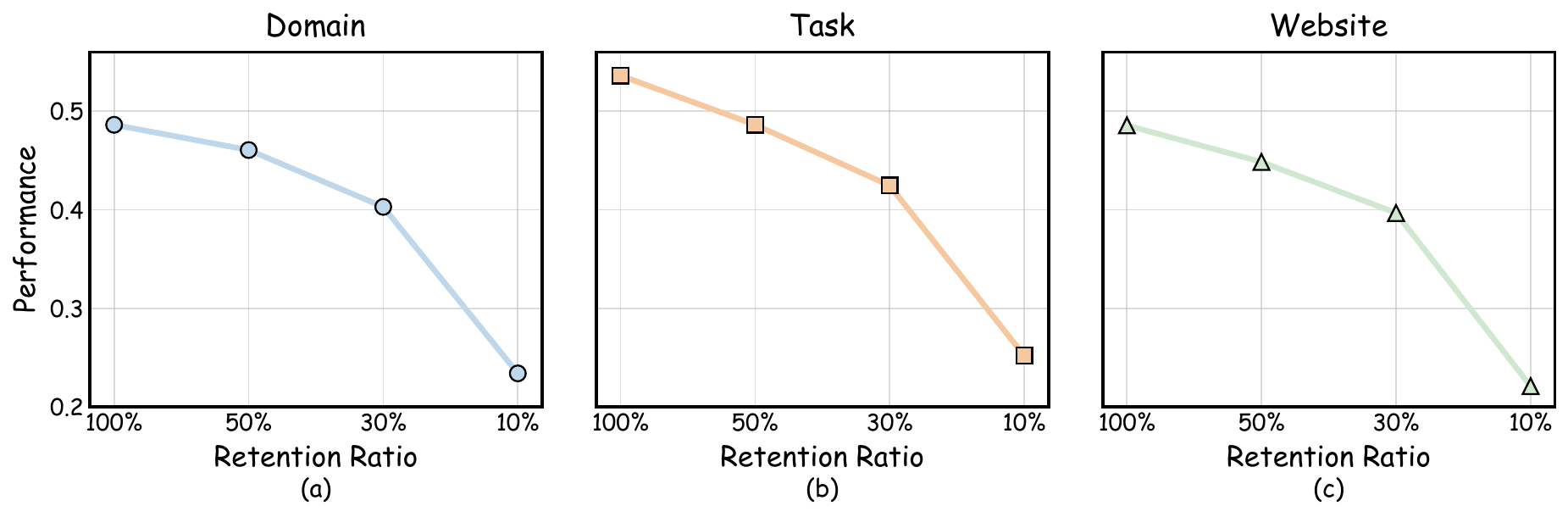}
    \caption{Performance of FastV when applied to the current screenshot
on Mind2Web under different visual-token retention ratios}
    \label{fig:curren_screenshot_mind2web}
\end{figure}

Throughout the main experiments, we apply token pruning only to 
historical screenshots while keeping the current screenshot intact. 
To validate this choice, we further apply FastV to the visual 
tokens of the current screenshot on AITW and Mind2Web using Qwen3-VL-2B. The results are reported in Table~\ref{tab:current_screenshot_pruning} and Figure~\ref{fig:curren_screenshot_mind2web}.

Pruning the current screenshot leads to substantial and monotonic
performance degradation on both AITW and Mind2Web. On AITW, even
when retaining 50\% of the visual tokens in the current screenshot,
the average score decreases from 0.7035 to 0.6612, corresponding to
an absolute drop of 4.23 percentage points. As the retention ratio is
reduced to 40\%, 30\%, and 20\%, the average score further decreases
to 0.6381, 0.6081, and 0.5461, respectively. A consistent trend is
observed on Mind2Web. The average score across the cross-domain,
cross-task, and cross-website settings decreases from 0.5028 to
0.4652 when 50\% of the visual tokens are retained, and further drops
to 0.4082 and 0.2357 at retention ratios of 30\% and 10\%,
respectively.

These results indicate that the current screenshot provides the most 
direct and indispensable visual evidence for predicting the next action, 
including recognizing the current interface state and locating the 
interaction target. In contrast, historical screenshots mainly provide 
auxiliary contextual information and therefore exhibit greater visual-token redundancy. Based on this observation, our main experiments focus on pruning visual tokens from historical screenshots while leaving the 
current screenshot unchanged.

\section{Composing ViT Feature Caching with LLM-side Token Pruning}
\label{app:cache_pruning}
\raggedbottom

ViT feature caching and LLM-side token pruning operate at different
stages of the inference pipeline and can therefore be composed. ViT
feature caching avoids redundant visual encoding when a screenshot
subsequently reappears in the history window, whereas LLM-side pruning
reduces the cost of processing redundant historical visual tokens.

For Qwen3-VL, caching only the final visual features is insufficient
because the model additionally uses DeepStack. Specifically, the vision
encoder produces three intermediate DeepStack features after vision
blocks 8, 16, and 24. These features are injected into the hidden states
after decoder blocks 0, 1, and 2, respectively. We therefore cache the
final merged visual features and all three merged DeepStack features for
each screenshot. All block indices in this section are zero-based.

Our pruning mask is constructed at the fourth decoder block (index 3)
and physically applied after this block, before the fifth decoder block
(index 4). Hence, all three DeepStack features have already been injected
before historical visual tokens are removed. The cache always retains
the complete visual representation of each resident screenshot, whereas
token pruning is applied only to the assembled sequence at the current
inference step. The joint inference procedure is summarized in
Algorithm~\ref{alg:joint_cache_pruning}.

We evaluate this composition on AITW using Qwen3-VL-2B. Random-0.1
retains 10\% of the visual tokens from each historical screenshot while
leaving the current screenshot intact. We compare four settings: the
original inference pipeline without caching, ViT feature caching alone,
Random-0.1 alone, and their combination.

As shown in Table~\ref{tab:cache_pruning}, ViT feature caching reduces
encoder latency from 212.08 to 69.82 ms per step, while Random-0.1
reduces LLM prefill latency from 193.05 to 103.67 ms per step. When
combined, the encoder latency remains low at 70.04 ms per step, and the
LLM prefill latency remains close to that of Random-0.1 at 105.69 ms per
step. The combined setting attains an average score of 0.6853, closely
matching the 0.6842 achieved by Random-0.1 alone. This result indicates
that feature caching introduces no additional performance degradation
beyond LLM-side pruning.

The combined strategy further reduces the end-to-end wall-clock time to
89.22 minutes, saving 6.48 minutes relative to ViT feature caching alone
and 11.15 minutes relative to Random-0.1 alone. These results demonstrate
that ViT feature caching is compatible with subsequent LLM-side token
pruning and that the two techniques can be jointly applied for further
end-to-end acceleration.

\begin{table*}[t]
\centering
\scriptsize
\setlength{\tabcolsep}{3pt}
\renewcommand{\arraystretch}{1.08}
\begin{tabular*}{\textwidth}{@{\extracolsep{\fill}}lccccccccc@{}}
\toprule
\textbf{Strategy}
& \textbf{General}
& \textbf{Single}
& \textbf{Webshopping}
& \textbf{Install}
& \textbf{GoogleApps}
& \textbf{Avg.}
& \shortstack{\textbf{Wall-clock}\\\textbf{(min)}}
& \shortstack{\textbf{Encoder}\\\textbf{(ms/step)}}
& \shortstack{\textbf{LLM Prefill}\\\textbf{(ms/step)}} \\
\midrule
No Cache
& 0.6496 & 0.7844 & 0.6579 & 0.7366 & 0.6891 & 0.7035
& 108.54 & 212.08 & 193.05 \\
ViT Feature Cache
& 0.6389 & 0.7784 & 0.6614 & 0.7310 & 0.7049 & 0.7029
& 95.70 & 69.82 & 202.54 \\
Random-0.1
& 0.6140 & 0.7605 & 0.6274 & 0.7223 & 0.6970 & 0.6842
& 100.37 & 192.08 & 103.67 \\
ViT + Random-0.1
& 0.6152 & 0.7574 & 0.6280 & 0.7287 & 0.6970 & 0.6853
& 89.22 & 70.04 & 105.69 \\
\bottomrule
\end{tabular*}
\caption{Compatibility of ViT feature caching and LLM-side random
token pruning on AITW using Qwen3-VL-2B. Random-0.1 retains 10\% of
the visual tokens from each historical screenshot. The combined
strategy preserves the performance and LLM prefill speed of
LLM-side pruning while additionally avoiding redundant visual
encoding.}
\label{tab:cache_pruning}
\end{table*}

\begin{algorithm}[H]
\caption{Joint ViT feature caching and LLM-side token pruning.}
\label{alg:joint_cache_pruning}
\footnotesize
\begin{algorithmic}[1]
\REQUIRE Encoder $E$; decoder blocks $\{M_\ell\}_{\ell=0}^{L-1}$;
instruction $G$; history size $\tau$; retention ratio $R$; rule $\mathcal{P}$
\ENSURE Action sequence for one episode
\STATE Initialize $\mathcal{C}\gets\emptyset$ and
$\mathcal{H}_0\gets\emptyset$
\FOR{each timestep $t$}
    \STATE Receive screenshot $o_t$ with identifier $k_t$
    \IF{$k_t\notin\mathcal{C}$}
        \STATE $(F_t,D_t,g_t)\gets E(o_t)$
        \STATE $\mathcal{C}[k_t]\gets(F_t,D_t,g_t)$
    \ENDIF
    \STATE $(h,V_H,V_C,D,p,B)\gets
    \operatorname{Assemble}(G,\mathcal{H}_t,k_t,\mathcal{C})$
    \STATE $V\gets V_H\cup V_C$
    \FOR{$\ell\gets0$ \textbf{to} $2$}
        \STATE $h\gets M_\ell(h;p,B)$
        \STATE $h[V]\gets h[V]+D^{(\ell)}$
    \ENDFOR
    \STATE $(h,S_3)\gets M_3(h;p,B)$
    \STATE $m\gets\mathcal{P}(S_3,V_H,R)$
    \STATE $(h,p,B)\gets\operatorname{Compact}(h,p,B,m)$
    \FOR{$\ell\gets4$ \textbf{to} $L-1$}
        \STATE $h\gets M_\ell(h;p,B)$
    \ENDFOR
    \STATE $a_t\gets\operatorname{Decode}(h;p,B)$
    \STATE $\mathcal{H}_{t+1}\gets
    \operatorname{Tail}_{\tau}(\mathcal{H}_t\mathbin{\|}[(k_t,a_t)])$
    \STATE $\mathcal{C}\gets
    \operatorname{Retain}(\mathcal{C},\operatorname{Keys}(\mathcal{H}_{t+1}))$
\ENDFOR
\STATE Clear $\mathcal{C}$ at the end of the episode
\end{algorithmic}
\end{algorithm}

\paragraph{Notation.}
At the beginning of timestep $t$, the ordered history window is
$\mathcal{H}_t=[(k_j,a_j)]_{j=\max(0,t-\tau)}^{t-1}$, where $k_j$
identifies screenshot $o_j$ and $a_j$ is the corresponding action. The
episode-local cache stores
$\mathcal{C}[k_j]=(F_j,D_j,g_j)$, where $F_j$ is the final merged visual
representation, $D_j=(D_j^{(0)},D_j^{(1)},D_j^{(2)})$ contains the three
merged DeepStack features, and $g_j$ contains the visual-token grid
metadata required for input assembly.

The $\operatorname{Assemble}$ operation retrieves the cached features
for the historical and current screenshots and interleaves them with
$G$ and the historical actions. It returns the initial hidden-state
sequence $h$; the historical and current visual-token index sets $V_H$
and $V_C$; the assembled DeepStack features
$D=(D^{(0)},D^{(1)},D^{(2)})$; the multimodal position indices $p$; and
the attention mask $B$. We write $M_\ell(h;p,B)$ for decoder block
$\ell$ applied with $p$ and $B$.

The pruning rule $\mathcal{P}$ returns a full-sequence keep mask $m$.
Instruction tokens, action tokens, and current visual tokens are always
retained, while an $R$ fraction of the tokens from each historical
screenshot is selected. The optional quantity $S_3$ denotes selection
statistics from the fourth decoder block; Random pruning does not use
$S_3$. The $\operatorname{Compact}$ operation applies $m$ consistently
to $h$, $p$, and the corresponding dimensions of $B$. It therefore
preserves the original position indices of retained tokens. Since all
three DeepStack features have already been injected, $D$ does not need
to be compacted. Neither selection nor compaction modifies the complete
features stored in $\mathcal{C}$; cache entries outside the updated
history window are evicted only to bound memory usage.

\flushbottom

\end{document}